\documentclass[10pt,letterpaper]{article}
\usepackage{alibaba-report}
\usepackage{colortbl}
\usepackage{pifont}
\usepackage{adjustbox}
\usepackage{needspace}
\usepackage{afterpage}
\usepackage{flafter}
\usepackage{wrapfig}
\makeatletter
\newcommand{\FinishSideFigure}{%
  \par
  \begingroup
    \setlength{\parskip}{0pt}%
    \global\advance\c@WF@wrappedlines-2\relax
    \@whilenum\c@WF@wrappedlines>\@ne\do{\noindent\strut\par}%
  \endgroup
  \WFclear}
\makeatother
\DeclareMicrotypeSet{paperprotrusion}{encoding={T1,OT1},family={lmr,cmr}}
\UseMicrotypeSet[protrusion]{paperprotrusion}
\newcommand{\paperfit}[1]{\adjustbox{max width=\linewidth}{#1}}
\newcommand{\heatcell}[2]{\cellcolor[HTML]{#1}{#2}}
\definecolor{LightCyan}{rgb}{0.94,1.0,1.0}
\definecolor{ForestGreen}{rgb}{0.13,0.55,0.13}
\definecolor{LightBlue}{rgb}{0.678,0.847,0.902}
\definecolor{DarkRed}{rgb}{0.7,0.1,0.1}
\newcommand{\algcomment}[1]{\State \textcolor{ForestGreen}{\textit{// #1}}}
\tcbset{aibox/.style={
  width=\linewidth,top=8pt,bottom=4pt,
  colback=blue!6!white,colframe=black,colbacktitle=black,
  enhanced,center,
  attach boxed title to top left={yshift=-0.1in,xshift=0.15in},
  boxed title style={boxrule=0pt,colframe=white}}}
\newtcolorbox{AIbox}[2][]{aibox,title=#2,#1}

\title{Perceive-to-Reason: Decoupling Perception and Reasoning for Fine-Grained Visual Reasoning}
\author{%
  Hongxing Li\textsuperscript{1,2,*}, Xiufeng Huang\textsuperscript{2,*},
  Dingming Li\textsuperscript{1}, Wenjing Jiang\textsuperscript{1,2}, Zixuan Wang\textsuperscript{1},\\
  Haolei Xu\textsuperscript{1,2}, Hanrong Zhang\textsuperscript{2},
  Haiwen Hong\textsuperscript{2,$\dagger$}, Longtao Huang\textsuperscript{2}, Hui Xue\textsuperscript{2},\\
  Weiming Lu\textsuperscript{1}, Jun Xiao\textsuperscript{1}, Yueting Zhuang\textsuperscript{1},
  Yongliang Shen\textsuperscript{1,$\dagger$}\\[6pt]
  {\reportaffiliationfont\mdseries\textsuperscript{1}Zhejiang University\qquad
  \textsuperscript{2}Yuvion Team, Alibaba Group}\\[4pt]
  {\normalfont\small\textsuperscript{*}Equal contribution\qquad
  \textsuperscript{$\dagger$}Corresponding authors}}
\newcommand{\shortpapertitle}{Perceive-to-Reason}
\newcommand{\codeurl}{https://github.com/ZJU-REAL/Perceive-to-Reason}
\newcommand{\modelsurl}{https://huggingface.co/hongxingli/P2R-4B}
\renewcommand{\paperlinks}{\par\vspace{3pt}\begingroup
  \sffamily\fontsize{7.5}{10}\selectfont
  \noindent\makebox[\linewidth][c]{%
    \scalebox{.9}{\paperlinkentry{\codeurl}{Code}}\hspace{24pt}%
    \scalebox{.9}{\paperlinkentry{\modelsurl}{Models}}}
  \par\endgroup}
\renewcommand{\authorcontributions}{}
\draftfalse
\AtBeginDocument{\hypersetup{
  pdftitle={Perceive-to-Reason: Decoupling Perception and Reasoning for Fine-Grained Visual Reasoning},
  pdfauthor={Hongxing Li, Xiufeng Huang, Dingming Li, Wenjing Jiang, Zixuan Wang, Haolei Xu, Hanrong Zhang, Haiwen Hong, Longtao Huang, Hui Xue, Weiming Lu, Jun Xiao, Yueting Zhuang, Yongliang Shen}}}

\begin{document}
\maketitle
\begin{abstract}
Fine-grained visual reasoning remains challenging for vision-language models, especially when small but critical visual cues are buried in high-resolution images. Existing approaches rely on repeated cropping or test-time visual search to introduce local evidence, but they typically do not explicitly distinguish perception from reasoning. In this paper, we propose \textbf{P}erceive-\textbf{to}-\textbf{R}eason (\textbf{P2R}), a unified framework that formulates fine-grained visual reasoning as a two-stage process: the model first localizes question-relevant evidence as a \emph{Perceiver}, and then answers the question as a \emph{Reasoner} based on the annotated image and cropped regions. To better align training with this decoupled formulation, we further introduce \textbf{P}erception-\textbf{R}easoning \textbf{A}lternating \textbf{GRPO} (\textbf{PRA-GRPO}), a role-aware reinforcement learning strategy that alternates between perception-focused and reasoning-focused updates using only final-answer supervision. Built on top of Qwen3-VL-Instruct-2B/4B/8B, P2R consistently improves performance across model scales. In particular, P2R-4B achieves 93.2\% on V-Star, 81.9\% on HR-Bench-4K, and 80.5\% on HR-Bench-8K, substantially outperforming its corresponding backbone. Further experiments show that the benefits of P2R extend beyond high-resolution benchmarks to broader multimodal reasoning tasks. These results suggest that explicitly decoupling perception from reasoning provides an effective framework for fine-grained visual reasoning.

\paperlinks
\end{abstract}
% Place and number the overview on page 3, after the preceding motivation figure.
\afterpage{\afterpage{\begin{figure}[H]
    \centering
    \includegraphics[width=\linewidth, trim=0 0.2cm 0 0, clip]{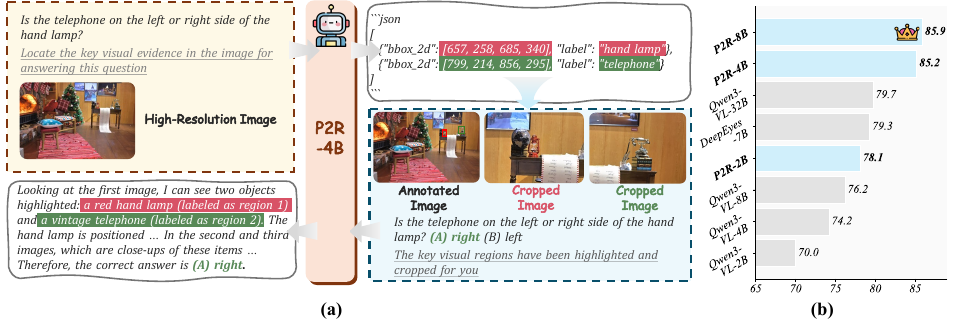}
    \caption{Overview of P2R. (a) Illustration of the proposed two-stage P2R inference pipeline. (b) Performance comparison on fine-grained visual reasoning benchmarks. P2R outperforms its base models across all scales.}

    \label{fig:teaser}
\end{figure}
}}
\section{Introduction}
Vision-language models (VLMs) have recently achieved strong performance on general visual understanding and reasoning tasks~\citep{huang2025vision-r1,yu2025perception}. Yet fine-grained visual reasoning remains challenging~\citep{wu2024v-star, wang2025hr-bench, zhang2024mme}, especially for tasks such as fine-grained text recognition and precise spatial relation understanding. Solving such tasks requires both locating subtle question-relevant evidence in high-resolution images and reasoning over it, that is, determining \emph{where to look} and \emph{how to reason}.

A simple diagnostic study suggests that perception is a major bottleneck in fine-grained visual reasoning. On V-Star~\citep{wu2024v-star}, Qwen3-VL-Instruct-4B~\citep{bai2025qwen3} improves from 81.7\% to 90.6\% when given oracle bounding boxes and cropped regions, indicating that many errors stem from failing to localize the right visual evidence (details in Appendix~\ref{app:preliminary}).

\Needspace{32\baselineskip}
\begin{wrapfigure}{r}{.45\textwidth}
\centering
\includegraphics[width=\linewidth]{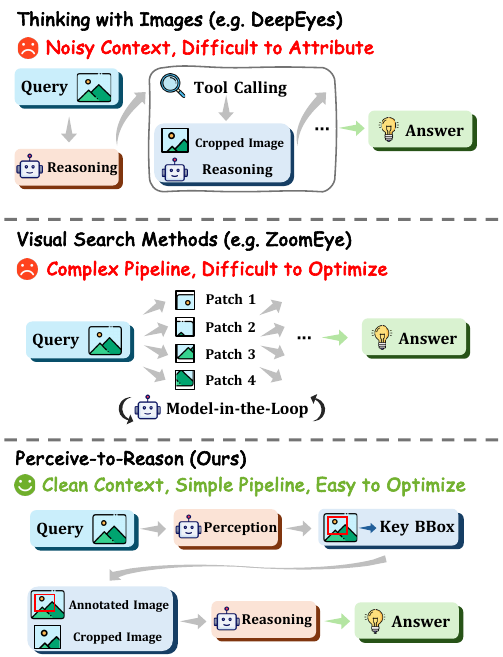}
\caption{Motivation of P2R. Prior methods inject local evidence via cropping or search without separating perception and reasoning. P2R instead adopts a decoupled perceive-to-reason paradigm.}

\label{fig:motivation}
\end{wrapfigure}

Existing approaches mainly address this challenge by injecting local evidence through region cropping or search~\citep{shao2024visual-cot, liu2024chain-of-spot}. They largely fall into two categories. \emph{Thinking with Images} methods~\citep{zheng2025deepeyes, wang2025pixelreasoner} interleave region exploration with reasoning, often producing long and noisy contexts. Visual search methods~\citep{shen2025zoomeye,li2025dyfo} locate key regions through test-time search, but typically rely on complex pipelines that are difficult to optimize~\citep{li2026reliable, liu2025hide}. More importantly, these approaches either entangle evidence localization with reasoning or externalize perception into a separate search process, so the model is not directly optimized for \emph{where to look} decisions, as illustrated in Figure~\ref{fig:motivation}.

These observations motivate a two-stage formulation of fine-grained visual reasoning: first localize the relevant evidence, then reason over it. However, training such a decoupled process is difficult under answer-only supervision, since errors may arise from either poor localization or flawed reasoning, making credit assignment ambiguous.

To address this issue, we propose \textbf{P}erceive-\textbf{to}-\textbf{R}eason (\textbf{P2R}), a unified framework that explicitly decomposes fine-grained visual reasoning into perception and reasoning. At inference time, P2R first localizes question-relevant evidence as a \emph{Perceiver}, and then answers the question as a \emph{Reasoner} based on the annotated image and cropped regions. This formulation makes evidence localization an explicit intermediate step rather than an implicit byproduct of answer generation.

To train this decoupled formulation, we further propose \textbf{P}erception-\textbf{R}easoning \textbf{A}lternating \textbf{GRPO} (\textbf{PRA-GRPO}), a role-aware reinforcement learning strategy built upon GRPO~\citep{shao2024deepseekmath}. PRA-GRPO alternates between perception-focused and reasoning-focused optimization while keeping the other role fixed, thereby converting final-answer correctness into a more attributable training signal for the active stage. In this way, P2R improves both evidence localization and answer generation using only final-answer supervision, without requiring ground-truth bounding box annotations.

Built on top of Qwen3-VL-Instruct~\citep{bai2025qwen3}, P2R consistently improves over its base models across all scales. In particular, P2R-4B achieves 93.2\% on V-Star~\citep{wu2024v-star}, 81.9\% on HR-Bench-4K~\citep{wang2025hr-bench}, and 80.5\% on HR-Bench-8K~\citep{wang2025hr-bench}, with substantial gains over the corresponding backbone. Further experiments show that the benefits of P2R extend beyond high-resolution benchmarks to broader multimodal reasoning tasks. Our main contributions are summarized as follows:

\begin{itemize}
    \item We propose P2R, a unified framework for fine-grained visual reasoning that formulates the task as a two-stage perceive-to-reason process, explicitly decoupling evidence localization from answer generation.

    \item We introduce PRA-GRPO, a role-aware reinforcement learning strategy that aligns training with the decoupled perceive-to-reason formulation, using only final-answer supervision without requiring bounding box annotations.

    \item Built on top of Qwen3-VL-Instruct models, P2R consistently delivers substantial gains across model scales and achieves state-of-the-art results on high-resolution fine-grained visual reasoning benchmarks.
\end{itemize}

\section{Related Work}
\paragraph{Fine-Grained Visual Reasoning}
Fine-grained visual reasoning requires models to identify subtle visual evidence and reason over it, and remains challenging for current VLMs~\citep{wu2024v-star, wang2025hr-bench, zhang2024mme, wei2026zooming, wang2025grasp, li2025viewspatial}. Existing methods mainly tackle this challenge by localizing key regions. One line of work follows the \emph{Thinking with Images} paradigm~\citep{hong2025deepeyesv2, lai2025mini-o3, fan2025grit, wang2025adatooler, zhao2025pyvision, zhao2026pyvision-rl}, where models iteratively zoom into relevant regions or invoke visual tools for interleaved visual-textual reasoning. For example, DeepEyes~\citep{zheng2025deepeyes} leverages reinforcement learning to improve visual tool use. Another line of work adopts visual search or test-time scaling~\citep{yu2025zoom-refine, shao2024visual-cot, khayatkhoei2025mllms, hu2024visual-ske} to identify informative subregions during inference; for instance, ZoomEye~\citep{shen2025zoomeye} performs hierarchical search over zoomed-in regions. However, these methods either entangle perception and reasoning within a single process or rely on external pipelines, without explicitly formulating fine-grained visual reasoning as a perceive-to-reason process. Our method, in contrast, explicitly decomposes the task and aligns training accordingly.

\paragraph{Reinforcement Learning in VLMs}
Recent studies have extended reinforcement learning (RL) from LLMs to VLMs, leading to notable progress in visual reasoning~\citep{yu2025perception, liu2025visual-rft, yang2025r1-onevision, li2025spatialladder, wang2025omniear, liu2025vlm-fo1, liu2025seg-zero, chen2025perception-before-reasoning, wang2026vl-rethinker}. Representative works such as Vision-R1~\citep{huang2025vision-r1} and MM-Eureka~\citep{meng2025mm} show that RL can significantly improve reasoning capabilities in VLMs, especially for visual mathematical reasoning. Perception-oriented methods such as VLM-R1~\citep{shen2025vlm-r1} and Perception-R1~\citep{yu2025perception} use rewards based on IoU or F1 to improve grounding and counting. However, existing RL approaches optimize perception and reasoning in isolation, without addressing their coordination. Our PRA-GRPO instead optimizes both within a unified framework via alternating updates.

\Needspace{8\baselineskip}
\section{Methodology}
\begin{figure}[t]
    \centering
    \includegraphics[width=\linewidth, trim=0 0.1cm 0 0, clip]{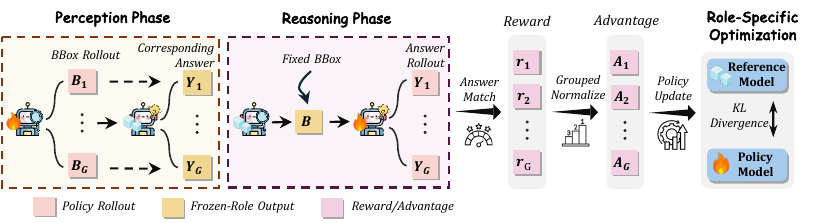}
    \caption{Overview of PRA-GRPO. Training alternates between a perception phase and a reasoning phase under shared model parameters. In each phase, the active role is optimized with GRPO while the other role is frozen, so that final answer correctness can be converted into a more attributable role-specific learning signal.}

    \label{fig:methodology}
\end{figure}

\subsection{P2R Framework Overview}
P2R is a unified framework that formulates fine-grained visual reasoning as a perceive-to-reason process. It consists of two tightly coupled components: a two-stage inference paradigm that explicitly separates evidence localization and answer generation, assigning these roles to a \emph{Perceiver} and a \emph{Reasoner}, and PRA-GRPO, a role-aware reinforcement learning strategy that aligns training with this decoupled formulation.

\subsection{Two-Stage P2R Inference}
Given an image-question pair $(I, Q)$, P2R structures fine-grained visual reasoning into two consecutive stages: perception and reasoning, as illustrated in Figure~\ref{fig:teaser}(a). We use a single underlying VLM with shared parameters $\theta$ throughout both stages. For notational convenience, we denote its role-conditioned behavior in the perception and reasoning stages as $\pi_p(\cdot;\theta)$ and $\pi_r(\cdot;\theta)$, corresponding to the \emph{Perceiver} and \emph{Reasoner}, respectively.

In the first stage, the model acts as a \emph{Perceiver} to make an explicit localization decision for the visual evidence most relevant to answering the question. Let $\tilde{Q}_p = \mathcal{T}_p(Q)$ denote a perception-oriented prompt derived from the original question. The \emph{Perceiver} predicts one or more bounding boxes as:
\begin{equation}
\mathcal{B} \sim \pi_p(\cdot \mid I, \tilde{Q}_p; \theta)
\end{equation}
where $\mathcal{B} = \{B_k\}_{k=1}^{K}$ denotes a set of rectangular regions in the image, and each $B_k = (x_1, y_1, x_2, y_2)$ specifies one region.

The predicted boxes are then transformed into two complementary visual inputs: an annotated image $I_a = \mathrm{annotate}(I, \mathcal{B})$ and cropped evidence images $I_c = \mathrm{crop}(I, \mathcal{B})$, where $\mathrm{annotate}(\cdot)$ overlays the predicted boxes on the original image and $\mathrm{crop}(\cdot)$ extracts the corresponding local regions.

In the second stage, the model acts as a \emph{Reasoner} and generates the final answer as:
\begin{equation}
Y \sim \pi_r(\cdot \mid I_a, I_c, Q; \theta)
\end{equation}

This two-stage formulation turns fine-grained visual reasoning into an explicitly structured process: the \emph{Perceiver} determines \emph{where to look}, while the \emph{Reasoner} focuses on \emph{how to reason} from the evidence. By making evidence localization an explicit intermediate step rather than an implicit byproduct of answer generation, P2R offers a more suitable formulation for fine-grained visual reasoning.

\subsection{PRA-GRPO}
\paragraph{Training the Decoupled Formulation}
While the two-stage P2R inference process explicitly separates perception from reasoning, training remains challenging because the final prediction depends on both stages. Incorrect evidence localization can mislead downstream reasoning, while correct evidence alone does not guarantee a correct answer if the subsequent reasoning is still flawed. As a result, it is difficult to improve fine-grained visual reasoning by treating the entire pipeline as a single undifferentiated optimization problem. Since supervision is only available at the level of the final answer, it is difficult to determine how the learning signal should be attributed across the two stages, especially when perception must be learned from downstream reasoning outcomes alone.

\paragraph{Role-Aware Alternating Optimization}
To address this issue, we propose PRA-GRPO, a role-aware reinforcement learning strategy that aligns optimization with the perceive-to-reason structure of P2R, as illustrated in Figure~\ref{fig:methodology}. The key idea is to convert final answer correctness into a more attributable training signal by alternating perception-focused and reasoning-focused updates. Intuitively, better evidence localization should increase the likelihood of successful downstream reasoning; therefore, even without ground-truth bounding box annotations, the final answer can serve as an indirect supervision signal for learning perception.

Concretely, PRA-GRPO alternates between optimizing the \emph{Perceiver} and the \emph{Reasoner}, while keeping the other role fixed. This turns final answer correctness into a role-aware supervision signal: in the perception phase, the quality of the predicted evidence is evaluated through the answer produced by a fixed Reasoner; in the reasoning phase, answer generation is optimized conditioned on evidence provided by a fixed Perceiver.

We now formalize this role-aware alternating optimization under the GRPO framework. Given an image-question-answer triplet $(I, Q, Y)$, we sample a group of $G$ rollouts from the role currently being optimized. In the perception phase, $o_i$ is a set of bounding boxes $\mathcal{B}_i \sim \pi_p(\cdot \mid I, \mathcal{T}_p(Q); \theta)$. Based on $\mathcal{B}_i$, we construct the annotated image $I_a^i$ and cropped evidence images $I_c^i$, which are then fed into a fixed \emph{Reasoner} to obtain the final answer $Y_i$. In the reasoning phase, $o_i$ is an answer $Y_i \sim \pi_r(\cdot \mid I_a, I_c, Q; \theta)$, where $(I_a, I_c)$ are constructed from the bounding boxes predicted by the fixed \emph{Perceiver}.

To keep supervision minimal and task-agnostic, we define a binary reward based solely on final-answer correctness:
\begin{equation}
r_i = \mathbb{I}[Y_i = Y]
\end{equation}
where $\mathbb{I}[\cdot]$ is the indicator function.

\paragraph{Role-Aware GRPO Objective}
We compute the group-relative advantage following GRPO~\citep{shao2024deepseekmath, guo2025deepseekr1}:
\begin{equation}
A_i = \frac{r_i - \mathrm{mean}(\{r_j\}_{j=1}^{G})}
{\mathrm{std}(\{r_j\}_{j=1}^{G}) + \epsilon}
\end{equation}
where $\epsilon$ is a small constant for numerical stability. We then optimize the active role in the current phase using the standard GRPO objective:
\begin{equation}
\begin{aligned}
\mathcal{J}_{\mathrm{GRPO}}^{\phi}(\theta)
=
\mathbb{E}_{x,\{o_i\}_{i=1}^{G}}
\Bigg[
\frac{1}{G}\sum_{i=1}^{G}
\min
\Big(
\rho_i A_i,\,
\\
\mathrm{clip}(\rho_i, 1\pm\varepsilon)A_i
\Big)
-\beta\,\mathrm{KL}\!\left[\pi_{\phi,\theta}\,\|\,\pi_{\mathrm{ref}}\right]
\Bigg]
\end{aligned}
\end{equation}
where
\begin{equation}
\rho_i = \frac{\pi_{\phi,\theta}(o_i \mid x)}{\pi_{\phi,\theta_{\mathrm{old}}}(o_i \mid x)}
\end{equation}
and $\phi \in \{p, r\}$ denotes the active role in the current optimization phase. Specifically, $x = (I, \mathcal{T}_p(Q))$ and $o_i = \mathcal{B}_i$ in the perception phase, while $x = (I_a, I_c, Q)$ and $o_i = Y_i$ in the reasoning phase.

\begin{table}[t!]
\centering
\small
\resizebox{\textwidth}{!}{
\begin{tabular}{lcccccccccc}
\toprule
\multirow{2}{*}{\textbf{Method}} & \multicolumn{3}{c}{\textbf{V-Star}} & \multicolumn{3}{c}{\textbf{HR-Bench-4K}} & \multicolumn{3}{c}{\textbf{HR-Bench-8K}} & \multirow{2}{*}{\textbf{Avg.}} \\
\cmidrule(lr){2-4} \cmidrule(lr){5-7} \cmidrule(lr){8-10}
 & \textbf{Attr} & \textbf{Spatial} & \textbf{Overall} & \textbf{FSP} & \textbf{FCP} & \textbf{Overall} & \textbf{FSP} & \textbf{FCP} & \textbf{Overall} & \\
\midrule
\rowcolor{gray!10}
\multicolumn{11}{l}{\textit{\textbf{Visual General Models}}} \\
GPT-4o~\citep{hurst2024gpt-4o} & - & - & 66.0 & 70.0 & 48.0 & 59.0 & 62.0 & 49.0 & 55.0 & 60.0 \\
o3 OpenAI~\citep{openai2025gpt-o3} & - & - & 95.7 & - & - & - & - & - & - & - \\
Qwen3-VL-Instruct-2B~\citep{bai2025qwen3} & 73.0 & 77.6 & 74.9 & 81.0 & 59.8 & 70.4 & 70.8 & 58.5 & 64.6 & 70.0 \\
Qwen3-VL-Instruct-4B~\citep{bai2025qwen3} & 79.1 & 85.5 & 81.7 & 81.8 & 66.0 & 73.8 & 73.8 & 60.3 & 67.0 & 74.2 \\
Qwen3-VL-Instruct-8B~\citep{bai2025qwen3} & 84.3 & 82.9 & 83.8 & 81.0 & 68.5 & 74.8 & 73.3 & 67.0 & 70.1 & 76.2 \\
Qwen3-VL-Instruct-32B~\citep{bai2025qwen3} & 86.1 & 88.2 & 86.9 & 86.0 & \underline{71.3} & 78.6 & 78.8 & 68.5 & 73.6 & 79.7 \\
\midrule
\rowcolor{gray!10}
\multicolumn{11}{l}{\textit{\textbf{Visual Search Methods}}} \\
DyFo-7B~\citep{li2025dyfo} & 80.0 & 82.9 & 81.2 & - & - & - & - & - & - & - \\
ZoomEye-7B~\citep{shen2025zoomeye} & \textbf{93.9} & 85.5 & 90.6 & 84.3 & 55.0 & 69.6 & 88.5 & 50.0 & 69.3 & 76.5 \\
\midrule
\rowcolor{gray!10}
\multicolumn{11}{l}{\textit{\textbf{Thinking with Images Methods}}} \\
DeepEyes-7B~\citep{zheng2025deepeyes} & 91.3 & 88.2 & 90.1 & 91.3 & 59.0 & 75.1 & 86.8 & 58.5 & 72.6 & 79.3 \\
PixelReasoner-7B~\citep{wang2025pixelreasoner} & 83.5 & 76.3 & 80.6 & 86.0 & 60.3 & 72.9 & 80.0 & 54.3 & 66.9 & 73.5 \\
Thyme-7B~\citep{zhang2025thyme} & 83.5 & 80.3 & 82.2 & 91.0 & 63.0 & 77.0 & 86.5 & 57.5 & 72.0 & 77.1 \\
\midrule
\rowcolor{gray!10}
\multicolumn{11}{l}{\textit{\textbf{P2R (Ours)}}} \\
P2R-2B & 84.3 & 84.2 & 84.3 & 89.8 & 60.1 & 75.1 & 86.8 & 62.7 & 74.8 & 78.1 \\
\textcolor{ForestGreen}{\quad \textit{$\Delta$ (vs Qwen3-VL-Instruct-2B)}} & \textcolor{ForestGreen}{\textit{+11.3}} & \textcolor{ForestGreen}{\textit{+6.6}} & \textcolor{ForestGreen}{\textit{+9.4}} & \textcolor{ForestGreen}{\textit{+8.8}} & \textcolor{ForestGreen}{\textit{+0.3}} & \textcolor{ForestGreen}{\textit{+4.7}} & \textcolor{ForestGreen}{\textit{+16.0}} & \textcolor{ForestGreen}{\textit{+4.2}} & \textcolor{ForestGreen}{\textit{+10.2}} & \textcolor{ForestGreen}{\textit{+8.1}} \\
P2R-4B & \underline{92.2} & \underline{94.7} & \underline{93.2} & \underline{92.3} & \textbf{71.5} & \textbf{81.9} & \underline{92.0} & \underline{69.0} & \underline{80.5} & \underline{85.2} \\
\textcolor{ForestGreen}{\quad \textit{$\Delta$ (vs Qwen3-VL-Instruct-4B)}} & \textcolor{ForestGreen}{\textit{+13.1}} & \textcolor{ForestGreen}{\textit{+9.2}} & \textcolor{ForestGreen}{\textit{+11.5}} & \textcolor{ForestGreen}{\textit{+10.5}} & \textcolor{ForestGreen}{\textit{+5.5}} & \textcolor{ForestGreen}{\textit{+8.1}} & \textcolor{ForestGreen}{\textit{+18.2}} & \textcolor{ForestGreen}{\textit{+8.7}} & \textcolor{ForestGreen}{\textit{+13.5}} & \textcolor{ForestGreen}{\textit{+11.0}} \\
P2R-8B & \underline{92.2} & \textbf{96.1} & \textbf{93.7} & \textbf{93.5} & 69.5 & \underline{81.5} & \textbf{94.8} & \textbf{70.5} & \textbf{82.6} & \textbf{85.9} \\
\textcolor{ForestGreen}{\quad \textit{$\Delta$ (vs Qwen3-VL-Instruct-8B)}} & \textcolor{ForestGreen}{\textit{+7.9}} & \textcolor{ForestGreen}{\textit{+13.2}} & \textcolor{ForestGreen}{\textit{+9.9}} & \textcolor{ForestGreen}{\textit{+12.5}} & \textcolor{ForestGreen}{\textit{+1.0}} & \textcolor{ForestGreen}{\textit{+6.7}} & \textcolor{ForestGreen}{\textit{+21.5}} & \textcolor{ForestGreen}{\textit{+3.5}} & \textcolor{ForestGreen}{\textit{+12.5}} & \textcolor{ForestGreen}{\textit{+9.7}} \\

\bottomrule
\end{tabular}
}
\caption{Quantitative results on V-Star, HR-Bench-4K, and HR-Bench-8K benchmarks. \textbf{Bold} denotes the best and \underline{underline} denotes the second best.}

\label{tab:v_star_hr_bench}
\end{table}

As both roles are instantiated by the same underlying VLM with shared parameters $\theta$, PRA-GRPO improves both evidence localization and answer generation within a unified model. More importantly, it allows perception to be learned from downstream reasoning outcomes through final-answer supervision alone, without requiring ground-truth bounding boxes or task-specific dense rewards.

See Appendix~\ref{app:method_details} for more method details.

\section{Experiments}
\subsection{Experimental Setup}
\paragraph{Baselines and Benchmarks}
We compare P2R against three groups of representative baselines:
(1) general-purpose VLMs, including proprietary models such as GPT-4o~\citep{hurst2024gpt-4o} and o3~\citep{openai2025gpt-o3}, as well as open-source Qwen3-VL~\citep{bai2025qwen3} models of different sizes;
(2) visual search methods, including DyFo~\citep{li2025dyfo} and ZoomEye~\citep{shen2025zoomeye}; and
(3) thinking-with-images methods, including DeepEyes~\citep{zheng2025deepeyes}, PixelReasoner~\citep{wang2025pixelreasoner}, and Thyme~\citep{zhang2025thyme}.
Our primary evaluation targets high-resolution fine-grained visual reasoning benchmarks, including V-Star~\citep{wu2024v-star} and HR-Bench~\citep{wang2025hr-bench}, which require precise perception of subtle visual evidence followed by downstream reasoning. We further report results on MME-RealWorld-Lite~\citep{zhang2024mme} to assess whether the benefits of P2R extend to broader real-world multimodal reasoning scenarios.

\paragraph{Training Dataset}
We build a 10K training set by sampling 3K examples from DeepEyes~\citep{zheng2025deepeyes}, 3K from VisualProbe~\citep{lai2025mini-o3}, and 4K from ZwZ~\citep{wei2026zooming}.

\paragraph{Training Details}
We instantiate P2R on top of Qwen3-VL-Instruct~\citep{bai2025qwen3} models using GRPO~\citep{shao2024deepseekmath} on 4 H100 GPUs. In each alternating stage, the Perceiver phase and the Reasoner phase are each trained for one epoch. For each prompt, we sample 8 rollouts, and set the KL coefficient~\citep{kullback1951kl} to 0.01.

See Appendix ~\ref{app:training_details} and ~\ref{app:eval_details} for more details.

\begin{table}[t]
\centering
\small
\resizebox{\textwidth}{!}{%
\begin{tabular}{lcccccccccc}
\toprule
\multirow{2}{*}{\textbf{Method}} & \multirow{2}{*}{\textbf{Overall}} & \multicolumn{5}{c}{\textbf{Perception}} & \multicolumn{4}{c}{\textbf{Reasoning}} \\
\cmidrule(lr){3-7} \cmidrule(lr){8-11}
 & & \textbf{OCR} & \textbf{RS} & \textbf{DT} & \textbf{MO} & \textbf{AD} & \textbf{OCR} & \textbf{DT} & \textbf{MO} & \textbf{AD} \\
\midrule
Qwen3-VL-Instruct-2B~\citep{bai2025qwen3} & 47.3 & 85.2 & 38.7 & 74.0 & 32.9 & 39.1 & 69.0 & 67.0 & 40.7 & 30.8 \\
Qwen3-VL-Instruct-4B~\citep{bai2025qwen3} & 47.7 & 88.0 & 38.0 & 80.0 & 34.8 & 33.7 & 74.0 & 77.0 & 34.7 & 31.8 \\
Qwen3-VL-Instruct-8B~\citep{bai2025qwen3} & 50.4 & 88.8 & 49.3 & 85.0 & 36.4 & 33.1 & \underline{81.0} & 79.0 & 38.7 & 34.3 \\
Qwen3-VL-Instruct-32B~\citep{bai2025qwen3} & 52.3 & 86.8 & 50.0 & \textbf{89.0} & 38.6 & 36.0 & \textbf{82.0} & \textbf{88.0} & 44.0 & 34.5 \\
DeepEyes-7B~\citep{zheng2025deepeyes} & 53.2 & \underline{90.0} & \underline{52.7} & \textbf{89.0} & 43.3 & 33.4 & 76.0 & 69.0 & 44.0 & 35.0 \\
PixelReasoner-7B~\citep{wang2025pixelreasoner} & 49.7 & 89.6 & 52.0 & \underline{86.0} & 38.9 & 30.9 & 71.0 & 72.0 & \underline{46.0} & 32.5 \\
\midrule
\rowcolor{gray!10}
P2R-2B & 51.3 & 88.4 & 43.3 & 78.0 & 39.8 & \underline{40.6} & 76.0 & 68.0 & 44.0 & 35.5 \\
\rowcolor{gray!10}
\textcolor{ForestGreen}{\quad \textit{$\Delta$ (vs Qwen3-VL-Instruct-2B)}} & \textcolor{ForestGreen}{\textit{+4.0}} & \textcolor{ForestGreen}{\textit{+3.2}} & \textcolor{ForestGreen}{\textit{+4.6}} & \textcolor{ForestGreen}{\textit{+4.0}} & \textcolor{ForestGreen}{\textit{+6.9}} & \textcolor{ForestGreen}{\textit{+1.5}} & \textcolor{ForestGreen}{\textit{+7.0}} & \textcolor{ForestGreen}{\textit{+1.0}} & \textcolor{ForestGreen}{\textit{+3.3}} & \textcolor{ForestGreen}{\textit{+4.7}} \\
\rowcolor{gray!10}
P2R-4B & \underline{54.8} & 88.8 & 50.0 & \underline{86.0} & \textbf{46.1} & 39.1 & 79.0 & 78.0 & \textbf{46.7} & \textbf{39.5} \\
\rowcolor{gray!10}
\textcolor{ForestGreen}{\quad \textit{$\Delta$ (vs Qwen3-VL-Instruct-4B)}} & \textcolor{ForestGreen}{\textit{+7.1}} & \textcolor{ForestGreen}{\textit{+0.8}} & \textcolor{ForestGreen}{\textit{+12.0}} & \textcolor{ForestGreen}{\textit{+6.0}} & \textcolor{ForestGreen}{\textit{+11.3}} & \textcolor{ForestGreen}{\textit{+5.4}} & \textcolor{ForestGreen}{\textit{+5.0}} & \textcolor{ForestGreen}{\textit{+1.0}} & \textcolor{ForestGreen}{\textit{+12.0}} & \textcolor{ForestGreen}{\textit{+7.7}} \\
\rowcolor{gray!10}
P2R-8B & \textbf{57.4} & \textbf{92.4} & \textbf{58.7} & \textbf{89.0} & \underline{44.5} & \textbf{46.9} & \underline{81.0} & \underline{81.0} & \textbf{46.7} & \underline{39.0} \\
\rowcolor{gray!10}
\textcolor{ForestGreen}{\quad \textit{$\Delta$ (vs Qwen3-VL-Instruct-8B)}} & \textcolor{ForestGreen}{\textit{+7.0}} & \textcolor{ForestGreen}{\textit{+3.6}} & \textcolor{ForestGreen}{\textit{+9.4}} & \textcolor{ForestGreen}{\textit{+4.0}} & \textcolor{ForestGreen}{\textit{+8.1}} & \textcolor{ForestGreen}{\textit{+13.8}} & \textcolor{ForestGreen}{\textit{+0.0}} & \textcolor{ForestGreen}{\textit{+2.0}} & \textcolor{ForestGreen}{\textit{+8.0}} & \textcolor{ForestGreen}{\textit{+4.7}} \\
\bottomrule
\end{tabular}
}

\caption{Quantitative results on MME-RealWorld-Lite benchmark. \textbf{Bold} denotes the best and \underline{underline} denotes the second best among all methods.}

\label{tab:mme_realworld}
\end{table}

\subsection{Main Results}

\paragraph{High-Resolution Benchmarks}
Table~\ref{tab:v_star_hr_bench} reports the results on V-Star~\citep{wu2024v-star}, HR-Bench-4K~\citep{wang2025hr-bench}, and HR-Bench-8K~\citep{wang2025hr-bench}. P2R consistently outperforms its corresponding Qwen3-VL-Instruct baselines across all scales, indicating that the proposed perceive-to-reason formulation is effective for fine-grained visual reasoning. Averaged over the three benchmarks, P2R-2B, P2R-4B, and P2R-8B improve upon their Qwen3-VL-Instruct counterparts by 8.1\%, 11.0\%, and 9.7\%, respectively. The gains are especially pronounced on HR-Bench-8K, highlighting the advantage of P2R in challenging high-resolution settings. P2R also compares favorably with prior visual search and thinking-with-images methods, with P2R-8B achieving the best average performance among all open-source models. A summary comparison is also shown in Figure~\ref{fig:teaser} (b).

\paragraph{General Perception and Reasoning Benchmark}
Table~\ref{tab:mme_realworld} reports the results on MME-RealWorld-Lite~\citep{zhang2024mme}, a broad benchmark covering diverse real-world multimodal perception and reasoning tasks. P2R consistently improves over its Qwen3-VL-Instruct backbones, with overall gains of 4.0\%, 7.1\%, and 7.0\% for the 2B, 4B, and 8B models, respectively. Notably, these improvements are broad rather than concentrated in a few categories: P2R improves performance across nearly all sub-tasks in both perception and reasoning. This suggests that the benefits of the perceive-to-reason formulation extend beyond high-resolution fine-grained settings to more general multimodal understanding scenarios. P2R-8B achieves the best overall performance among all compared methods.

\begin{figure}[H]
  \centering
  \begin{minipage}[t]{.48\textwidth}
    \centering
    \captionsetup{justification=raggedright,singlelinecheck=false}
    \includegraphics[height=43mm]{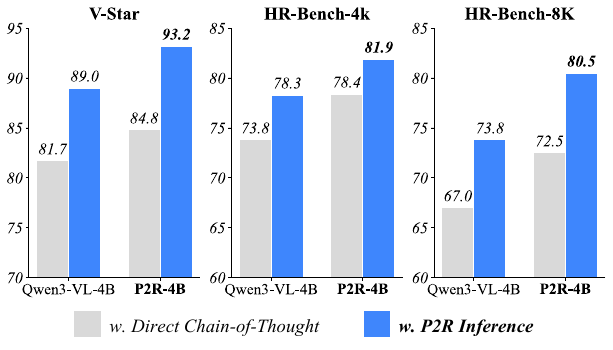}
    \caption{Direct CoT versus P2R inference on Qwen3-VL-Instruct-4B and P2R-4B across three high-resolution benchmarks.}
    \label{fig:ablation_p2r_inference}
  \end{minipage}\hspace{6pt}%
  \begin{minipage}[t]{.48\textwidth}
    \centering
    \captionsetup{justification=raggedright,singlelinecheck=false}
    \includegraphics[height=43mm]{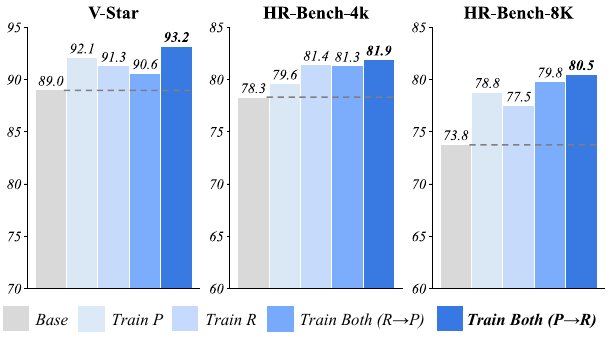}
    \caption{PRA-GRPO training ablations on Qwen3-VL-Instruct-4B. The dashed line marks the baseline with P2R inference.}
    \label{fig:ablation_training_component}
  \end{minipage}
\end{figure}

\subsection{Ablation Study}

\paragraph{Effect of P2R Inference}
Figure~\ref{fig:ablation_p2r_inference} compares direct chain-of-thought~\citep{wei2022chain-of-thought} prompting with the proposed P2R inference on both Qwen3-VL-Instruct-4B and P2R-4B. P2R inference consistently improves performance across all three benchmarks for both models, suggesting that the perceive-to-reason decomposition is beneficial already at inference time. On V-Star, for example, replacing direct CoT with P2R inference improves the score from 81.7\% to 89.0\% for Qwen3-VL-Instruct-4B, and from 84.8\% to 93.2\% for P2R-4B. Moreover, P2R-4B remains stronger than Qwen3-VL-Instruct-4B even under direct CoT prompting, indicating that the benefits of PRA-GRPO are not limited to the dedicated P2R inference pipeline. Combining P2R training with P2R inference yields the strongest performance on all three benchmarks.

% Keep this paired float at the top of the next page, above the analysis text.
\begin{figure}[!t]
  \centering
  \begin{minipage}[b]{.48\textwidth}
    \centering
    \captionsetup{justification=raggedright,singlelinecheck=false}
    \begin{minipage}[c][56mm][c]{\linewidth}
      \includegraphics[width=\linewidth]{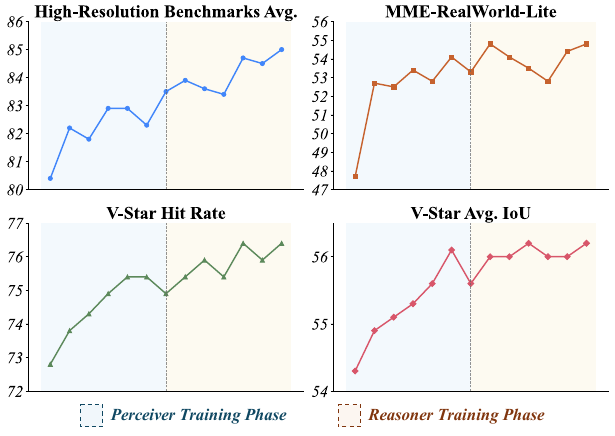}
    \end{minipage}
    \caption{Training dynamics of PRA-GRPO during the Perceiver and Reasoner phases, tracking localization and benchmark performance.}
    \label{fig:training_dynamics}
  \end{minipage}\hspace{6pt}%
  \begin{minipage}[b]{.48\textwidth}
    \captionsetup{type=table,position=bottom,justification=raggedright,singlelinecheck=false}
    % Included in the right-hand panel beside the training-dynamics figure.
\centering
\begin{minipage}[c][56mm][c]{\linewidth}
\centering
\scriptsize
\paperfit{
\renewcommand{\arraystretch}{2.15}
\setlength{\tabcolsep}{3pt}

\begin{tabular}{l|l|cccc}
\hline
\rowcolor{gray!20}
\textbf{P Ckpt} & \textbf{R Ckpt} & \textbf{V-Star} & \textbf{HR-4K} & \textbf{HR-8K} & \textbf{Avg.} \\
\hline\hline

\textit{Base} & \textit{Base}
& \heatcell{F7FCF5}{89.0}
& \heatcell{F7FCF5}{78.3}
& \heatcell{F7FCF5}{73.8}
& \heatcell{F7FCF5}{80.4} \\
\hline

\textit{P-Only} & \textit{Base}
& \heatcell{E5F5E0}{90.6}
& \heatcell{F1FAEE}{78.6}
& \heatcell{C7E9C0}{78.1}
& \heatcell{D9F0D3}{82.4} \\

\textit{P-Only} & \textit{P-Only}
& \heatcell{74C476}{92.1}
& \heatcell{BAE4B3}{79.6}
& \heatcell{A1D99B}{78.8}
& \heatcell{A1D99B}{83.5} \\
\hline

\textit{Base} & \textit{R-Only}
& \heatcell{D9F0D3}{90.8}
& \heatcell{98D594}{80.5}
& \heatcell{E5F5E0}{76.9}
& \heatcell{C7E9C0}{82.7} \\

\textit{R-Only} & \textit{R-Only}
& \heatcell{C7E9C0}{91.3}
& \heatcell{74C476}{81.4}
& \heatcell{D9F0D3}{77.5}
& \heatcell{74C476}{83.4} \\
\hline

\textit{P-Only} & \textit{R-Only}
& \heatcell{74C476}{92.1}
& \heatcell{74C476}{81.4}
& \heatcell{00441B}{\textcolor{white}{81.0}}
& \heatcell{238B45}{\textcolor{white}{84.8}} \\

\textit{P2R-Full} & \textit{P2R-Full}
& \heatcell{00441B}{\textcolor{white}{93.2}}
& \heatcell{00441B}{\textcolor{white}{81.9}}
& \heatcell{006D2C}{\textcolor{white}{80.5}}
& \heatcell{00441B}{\textcolor{white}{85.2}} \\

\hline
\end{tabular}
}

\begin{minipage}{\linewidth}
\centering
\scriptsize
Worst\;
\colorbox[HTML]{F7FCF5}{\rule{0.6cm}{0pt}\rule{0pt}{0.13cm}}
\colorbox[HTML]{E5F5E0}{\rule{0.6cm}{0pt}\rule{0pt}{0.13cm}}
\colorbox[HTML]{C7E9C0}{\rule{0.6cm}{0pt}\rule{0pt}{0.13cm}}
\colorbox[HTML]{74C476}{\rule{0.6cm}{0pt}\rule{0pt}{0.13cm}}
\colorbox[HTML]{238B45}{\rule{0.6cm}{0pt}\rule{0pt}{0.13cm}}
\colorbox[HTML]{00441B}{\rule{0.6cm}{0pt}\rule{0pt}{0.13cm}}
\;Best
\end{minipage}
\end{minipage}

\caption{Shared-parameter ablations with base, role-specific, and fully trained checkpoints on three fine-grained visual reasoning benchmarks.}

\label{tab:shared_params_analysis}

  \end{minipage}
\end{figure}

\paragraph{Effect of PRA-GRPO Training}
Figure~\ref{fig:ablation_training_component} compares different PRA-GRPO training strategies on top of Qwen3-VL-Instruct-4B with P2R inference. \emph{Train P} and \emph{Train R} optimize only the perceiver or the reasoner, while \emph{Train Both} alternates between the two roles. Optimizing either role alone already improves over the no-training baseline, suggesting that both perception and reasoning can benefit from role-aware training. Alternating updates over both roles leads to further gains across benchmarks. On V-Star, the order $P \rightarrow R$ achieves 93.2\%, outperforming $R \rightarrow P$ at 90.6\%, suggesting that localizing evidence first better supports reasoning by providing more accurate visual inputs. These results support the effectiveness of aligning training with the decoupled perceive-to-reason formulation.

\subsection{Further Analysis}

\paragraph{Training Dynamics}
Figure~\ref{fig:training_dynamics} shows the training dynamics of PRA-GRPO. During the \emph{Perceiver} phase, both V-Star Hit Rate and Avg. IoU improve steadily, indicating more accurate localization of question-relevant evidence. We define Hit Rate as 1 if the center of a predicted box falls inside the ground-truth box, making it a simple proxy for localization. Performance on the high-resolution benchmark average and MME-RealWorld-Lite also improves in this phase. After switching to the \emph{Reasoner} phase, benchmark performance continues to increase while the localization metrics remain stable or improve slightly, suggesting complementary gains from the two roles. More training dynamics are provided in Appendix~\ref{app:additional_training_dynamics}.

\paragraph{Shared-Parameter Analysis}
To study the effect of parameter sharing, we initialize the \emph{Perceiver} and \emph{Reasoner} in P2R inference with different checkpoints, including the base model, the \emph{P-Only} and \emph{R-Only} checkpoints, and the final PRA-GRPO checkpoint, and evaluate their combinations on three fine-grained visual reasoning benchmarks.
Here, \emph{Base} denotes the original model, \emph{P-Only} and \emph{R-Only} are trained only for the perceiver or reasoner role, and \emph{P2R-Full} is the final checkpoint after full PRA-GRPO training.

Table~\ref{tab:shared_params_analysis} reveals clear cross-role transfer under shared parameters. Replacing the \emph{Base} \emph{Reasoner} with the \emph{P-Only} checkpoint improves the average score from 82.4\% to 83.5\% (\emph{P-Only + Base} vs.\ \emph{P-Only + P-Only}), indicating that \emph{Perceiver}-only training also benefits the model when reused in the \emph{Reasoner} role. Likewise, replacing the \emph{Base} \emph{Perceiver} with the \emph{R-Only} checkpoint increases the average score from 82.7\% to 83.4\% (\emph{Base + R-Only} vs.\ \emph{R-Only + R-Only}), suggesting that \emph{Reasoner}-only training also transfers to the \emph{Perceiver} role. However, simply combining separately trained checkpoints remains weaker than the final alternating model (84.8\% vs. 85.2\%), suggesting that PRA-GRPO better integrates both capabilities within a single shared model. In addition, this shared-parameter design is deployment-friendly, requiring only one model at inference time.

\paragraph{Generalization across Specific Perception and General Reasoning Tasks}
Table~\ref{tab:task_generalization} evaluates P2R-4B beyond the primary high-resolution benchmarks. ReasonSeg~\citep{lai2024lisa} tests query-conditioned grounding, while CountBench~\citep{paiss2023teaching} and CountQA~\citep{tamarapalli2025countqa} focus on visual counting. MMStar~\citep{chen2024we} and BabyVision~\citep{chen2026babyvision} assess broader multimodal understanding and visual logical reasoning. P2R-4B consistently outperforms Qwen3-VL-Instruct-4B, with gains of 1.1, 0.6, 2.0, 1.3, and 1.5 percentage points, respectively. The ReasonSeg improvement indicates transfer to explicit localization without bounding-box supervision. The counting gains further extend this transfer to tasks requiring evidence from multiple instances. Together with the improvements on MMStar and BabyVision, these results suggest that the benefits of P2R extend across both specific perception tasks and broader reasoning settings.

\begin{table}[H]
\centering
\small
\scalebox{0.92}{
\renewcommand{\arraystretch}{1.2}
\begin{tabular}{lccccc}
\toprule
\multirow{2}{*}{\textbf{Method}}
& \multicolumn{3}{c}{\textbf{Specific Perception}}
& \multicolumn{2}{c}{\textbf{General Reasoning}} \\
\cmidrule(lr){2-4} \cmidrule(lr){5-6}
& \textbf{ReasonSeg} & \textbf{CountBench} & \textbf{CountQA} & \textbf{MMStar} & \textbf{BabyVision} \\
\midrule
Qwen3-VL-4B & 61.6 & 84.9 & 28.1 & 69.7 & 13.7 \\
\midrule
\rowcolor{gray!10}
P2R-4B & \textbf{62.7} & \textbf{85.5} & \textbf{30.1} & \textbf{71.0} & \textbf{15.2} \\
\rowcolor{gray!10}
\textcolor{ForestGreen}{\textit{$\Delta$}}
& \textcolor{ForestGreen}{\textit{+1.1}}
& \textcolor{ForestGreen}{\textit{+0.6}}
& \textcolor{ForestGreen}{\textit{+2.0}}
& \textcolor{ForestGreen}{\textit{+1.3}}
& \textcolor{ForestGreen}{\textit{+1.5}} \\
\bottomrule
\end{tabular}
}
\caption{Generalization across specific perception and general reasoning benchmarks. ReasonSeg reports the average Acc@0.5 over the test and validation splits.}
\label{tab:task_generalization}
\end{table}

\Needspace{16\baselineskip}
\begingroup
\setlength{\intextsep}{4pt plus 1pt minus 1pt}
\begin{wrapfigure}{r}{.36\textwidth}
    \centering
    \includegraphics[width=\linewidth]{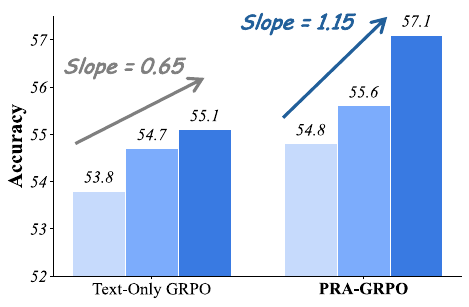}
    \caption{Accuracy over three training iterations for text-only GRPO and PRA-GRPO. PRA-GRPO achieves larger overall gains.}
    \label{fig:scaling_analysis}

\end{wrapfigure}
\noindent{\bfseries Training Scaling Analysis.}\enspace Figure~\ref{fig:scaling_analysis} shows the performance growth of PRA-GRPO and text-only GRPO over three training iterations on MME-RealWorld-Lite. PRA-GRPO rises from 54.8\% to 57.1\% with a fitted slope of 1.15, compared with 53.8\% to 55.1\% and a slope of 0.65 for text-only GRPO. Its advantage widens from 1.0 to 2.0 percentage points, with a larger gain in the final iteration (1.5 versus 0.4 points), suggesting that the benefit of role-aware training persists with repeated updates.

A plausible explanation is that the two roles reinforce each other: a stronger \emph{Perceiver} provides more relevant visual evidence, while a stronger \emph{Reasoner} can yield more informative answer-based feedback for optimizing the \emph{Perceiver}. This highlights an advantage of PRA-GRPO: alternating updates allow both evidence localization and evidence-conditioned reasoning to improve over successive training iterations.

\FinishSideFigure
\endgroup

\begin{figure}[H]
    \centering
    \includegraphics[width=\linewidth]{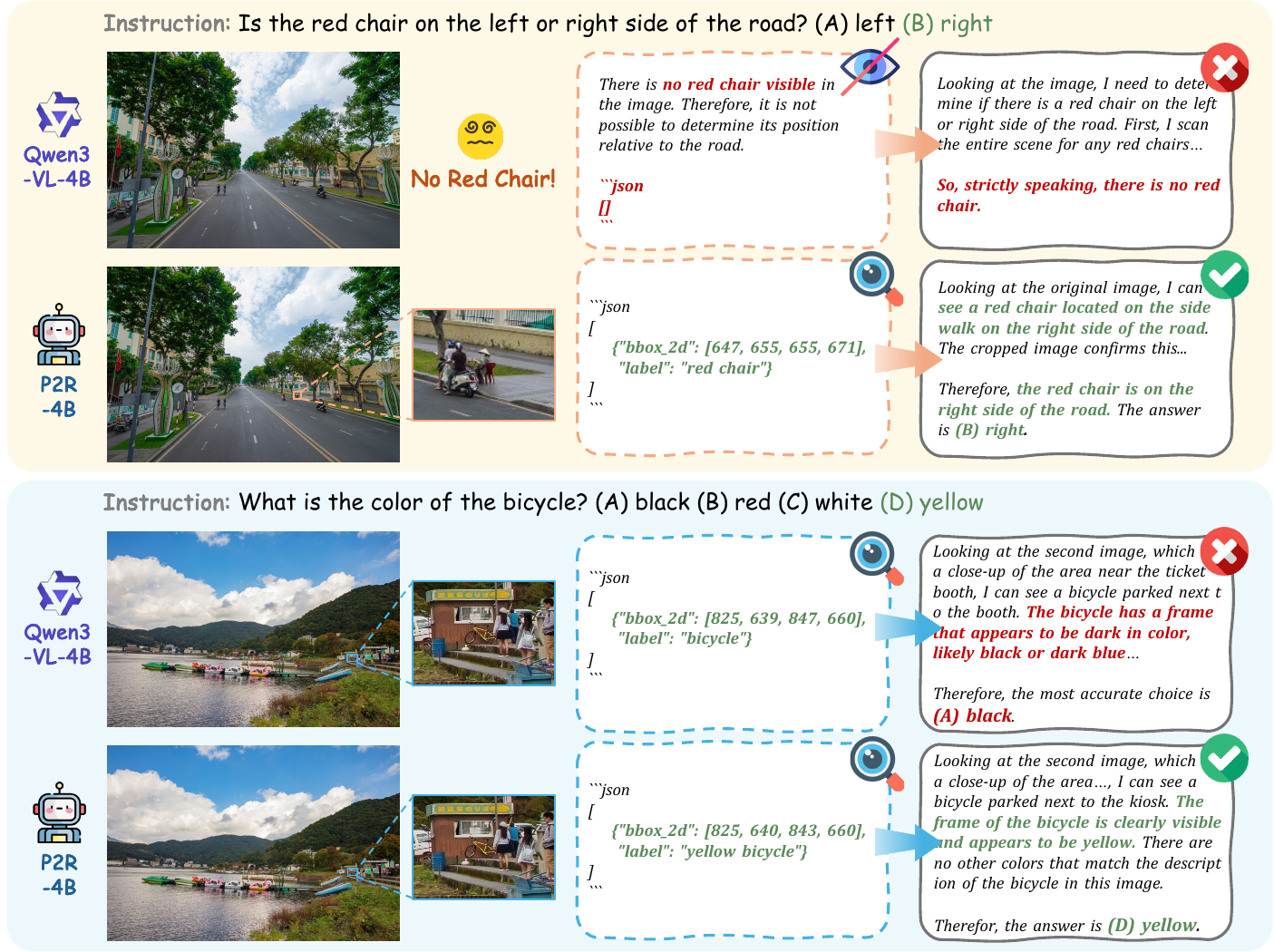}
    \caption{Representative examples from the V-Star benchmark, comparing Qwen3-VL-4B and P2R-4B.}
    \label{fig:case_study}
\end{figure}

\paragraph{Case Studies}
Figure~\ref{fig:case_study} presents two V-Star examples illustrating different sources of error. In the first, Qwen3-VL-Instruct-4B misses a small red chair and reports that it is absent, preventing a correct judgment of its spatial relation to the road. P2R-4B localizes the chair and uses the visual evidence to correctly identify it as lying on the right side. In the second, the baseline identifies the relevant bicycle near the kiosk but misrecognizes its frame as black. P2R-4B instead identifies the bicycle as yellow, illustrating that locating the target and interpreting its attributes are distinct requirements.

The two cases highlight the complementary roles of perception and reasoning: missing evidence can undermine the answer from the outset, while successful localization alone does not ensure correct use of fine-grained details. They provide qualitative support for explicitly separating evidence localization from evidence-conditioned answer generation.

See Appendix ~\ref{app:additional_analysis} for additional analysis.

\clearpage

\Needspace{8\baselineskip}
\section{Conclusion}
We propose P2R, a unified framework that decouples evidence localization from answer generation for fine-grained visual reasoning. Its training strategy, PRA-GRPO, alternately optimizes the Perceiver and Reasoner through shared parameters using only final-answer supervision, without ground-truth bounding boxes. Built on Qwen3-VL-Instruct models, P2R consistently improves performance on high-resolution benchmarks and general perception and reasoning tasks. These results highlight the value of explicitly separating perception and reasoning while aligning their optimization within a shared model.

\section{Limitations}
This work has several limitations. First, although P2R is simple and effective, its two-stage pipeline introduces additional inference cost compared with direct prompting. Second, due to limited computational resources, we have not explored PRA-GRPO at larger training scales, so its full scaling behavior remains unclear. Third, PRA-GRPO relies only on final-answer supervision, which avoids the need for bounding box annotations but also provides a sparse learning signal. Finally, our evaluation mainly focuses on fine-grained visual reasoning and related multimodal benchmarks; broader generalization to interactive or long-horizon settings remains for future work.

\bibliographystyle{plainnat}
\bibliography{references}

\clearpage
\appendix
\section{Diagnostic Study}
\label{app:preliminary}

We conduct a simple diagnostic study on V-Star~\citep{wu2024v-star} to probe a key question behind this work: is fine-grained visual reasoning limited more by reasoning or by perception? Using Qwen3-VL-4B, we compare a standard Direct CoT setting with an \emph{Oracle Hint} setting that provides ground-truth bounding boxes from the official V-Star annotation file, together with the corresponding cropped regions. Table~\ref{tab:diagnostic_study} shows that Oracle Hint improves overall accuracy from 81.7\% to 90.6\%, with consistent gains on both attribute and spatial questions. The result suggests that many failures are caused not by the inability to reason over evidence, but by the inability to first find the right evidence to reason over.

\begin{table}[H]
\centering
\small
\paperfit{
\begin{tabular}{lccc}
\toprule
\multirow{2}{*}{\textbf{Method}} & \multicolumn{3}{c}{\textbf{V-Star}} \\
\cmidrule(lr){2-4}
 & \textbf{Attr} & \textbf{Spatial} & \textbf{Overall} \\
\midrule
\rowcolor{gray!10}
\multicolumn{4}{l}{\textit{\textbf{Qwen3-VL-4B}}} \\
\quad + Direct CoT & 79.1 & 85.5 & 81.7 \\
\quad + Oracle Hint & 87.8 & 94.7 & 90.6 \\
\bottomrule
\end{tabular}
}
\caption{Diagnostic study on V-Star using Qwen3-VL-4B. \emph{Direct CoT} uses only the input image, while \emph{Oracle Hint} additionally provides ground-truth bounding boxes and cropped regions.}
\label{tab:diagnostic_study}
\end{table}

\section{Methodology Details}
\label{app:method_details}

\subsection{P2R Inference Details}
P2R inference consists of a Perceiver stage for localizing question-relevant evidence and a Reasoner stage for answering based on the localized evidence. We use the following prompts.

\begin{AIbox}{Prompt for Perceiver}
\small
    \textbf{System Prompt}: \textit{``You are a helpful assistant.''} \\
    \textbf{User Prompt}: \textit{\textbf{\{question\}} + ``Please carefully observe the image first to identify the object(s) referred to in the question. Note that each object type appears only once in the image. Then provide the 2D bounding box coordinates and labels of the related objects in JSON format.''}
\end{AIbox}

\begin{AIbox}{Prompt for Reasoner}
\small
    \textbf{System Prompt}: \textit{``You are a helpful assistant.''} \\
    \textbf{User Prompt}: \textit{\textbf{\{question\}} + ``The key visual regions have been highlighted and cropped for you. Think step by step.''}
\end{AIbox}

The predicted boxes are highlighted on the original image and cropped into local patches. Both the highlighted image and the local crops are then provided to the Reasoner.

\subsection{PRA-GRPO Details}
\label{app:pra_grpo_details}

Algorithm~\ref{alg:pragrpo} presents the training procedure of PRA-GRPO. Training alternates between Perceiver and Reasoner phases. In each phase, the active role is optimized while the other role is frozen using the checkpoint from the previous stage. For each training sample, we draw a group of rollouts from the active role, compute binary rewards from final-answer correctness, and derive group-relative advantages under GRPO. The resulting objective is then used to update only the active role.

The predicted bounding boxes are parsed from the model outputs using regular expressions. During post-processing, to keep the visual context within the input limit, we do not restrict the number of boxes rendered on the original image, but crop local patches from only the first three predicted boxes.

\begin{algorithm}[H]
\caption{PRA-GRPO}
\label{alg:pragrpo}
\small
\begin{algorithmic}[1]
\Require Perceiver $\pi_p$, Reasoner $\pi_r$, dataset $\mathcal{D}$, group size $G$, stage schedule
\For{each iteration}
    \State Sample a mini-batch $(I, Q, Y)$ from $\mathcal{D}$
    \State Select the active role $\phi \in \{p, r\}$ by the schedule
    \State Load the previous-stage checkpoint for the other role

    \algcomment{Step 1: Role-aware rollout}
    \For{each $(I, Q, Y)$ in the mini-batch}
        \If{$\phi = p$}
            \State Set $x \gets (I, \mathcal{T}_p(Q))$
            \State Sample $\{\mathcal{B}_i\}_{i=1}^{G} \sim \pi_p(\cdot \mid x)$
            \For{$i = 1, \dots, G$}
                \State Derive $(I_a^i, I_c^i)$ from $\mathcal{B}_i$
                \State Use the frozen Reasoner to obtain $Y_i$
                \State Set $o_i \gets \mathcal{B}_i$
            \EndFor
        \Else
            \State Obtain $\mathcal{B}$ from the frozen Perceiver
            \State Derive $(I_a, I_c)$ from $\mathcal{B}$
            \State Set $x \gets (I_a, I_c, Q)$
            \State Sample $\{Y_i\}_{i=1}^{G} \sim \pi_r(\cdot \mid x)$
            \For{$i = 1, \dots, G$}
                \State Set $o_i \gets Y_i$
            \EndFor
        \EndIf

        \algcomment{Step 2: Reward computation}
        \For{$i = 1, \dots, G$}
            \State Compute reward $r_i \gets \mathbb{I}[Y_i = Y]$
        \EndFor

        \algcomment{Step 3: Group-relative advantage}
        \State Compute $\mu \gets \mathrm{mean}(\{r_i\}),\ \sigma \gets \mathrm{std}(\{r_i\})$
        \For{$i = 1, \dots, G$}
            \State Compute $A_i \gets (r_i-\mu)/(\sigma+\epsilon)$
        \EndFor

        \algcomment{Step 4: GRPO policy update}
        \State Compute $\mathcal{L}_{\mathrm{GRPO}}$ from $\{o_i, A_i\}_{i=1}^{G}$
        \State Update the active role with $\mathcal{L}(\theta)=\mathcal{L}_{\mathrm{GRPO}}(\theta)$
    \EndFor
\EndFor
\end{algorithmic}
\end{algorithm}

\section{Training Details}
\label{app:training_details}

\subsection{Training Dataset}
\label{app:training_dataset}

We construct a 10K training set by random sampling from three complementary data sources: 3K examples from DeepEyes~\citep{zheng2025deepeyes}, 3K from VisualProbe~\citep{lai2025mini-o3}, and 4K from ZwZ~\citep{wei2026zooming}. These sources provide diverse supervision for fine-grained visual perception and evidence localization, while also covering different difficulty levels: DeepEyes is relatively easier, ZwZ presents medium-difficulty fine-grained perception cases, and VisualProbe is the most challenging due to small targets, cluttered scenes, and many distractors.

\begin{itemize}
    \item \textbf{DeepEyes:} We sample 3K examples from DeepEyes as a relatively easy source of training data. DeepEyes is curated for visually useful evidence and fine-grained perception, with filtering procedures for difficulty, answer validity, and perception utility. This makes it a suitable starting point for learning basic evidence localization behavior.

    \item \textbf{ZwZ:} We sample 4K examples from ZwZ as a medium-difficulty source of fine-grained perception data. ZwZ is synthetically generated by Region-to-Image distillation: strong teacher models first create question-answer pairs on micro-cropped regions, and the supervision is then distilled back to the full image with explicit region grounding. The resulting samples emphasize subtle local details while remaining more controlled than naturally hard search problems.

    \item \textbf{VisualProbe:} We sample 3K examples from VisualProbe as the hardest portion of the training mixture. Built from high-resolution images with very small targets and many distractors, it places strong demands on identifying sparse and localized visual evidence under clutter, making it particularly suitable for training robust perception behavior.
\end{itemize}

Overall, this mixture provides a coarse-to-hard spectrum of training difficulty, from relatively accessible grounding examples in DeepEyes, to medium-difficulty fine-grained cases in ZwZ, and finally to challenging visual search instances in VisualProbe. Despite using only 10K training examples in total, our method achieves significant performance gains, highlighting both the effectiveness of the proposed training framework and its strong data efficiency.

\begin{table}[H]
\centering
\paperfit{%
\begin{tabular}{lc}
\toprule
\textbf{Parameter} & \textbf{Value} \\
\midrule
algorithm.adv\_estimator & grpo \\
train\_batch\_size & 64 \\
truncation & error \\
filter\_overlong\_prompts & True \\
rollout.n & 8 \\
lr & $1\times10^{-6}$ \\
ppo\_mini\_batch\_size & 64 \\
ppo\_micro\_batch\_size\_per\_gpu & 8 \\
use\_kl\_loss & True \\
kl\_loss\_coef & $1\times10^{-2}$ \\
kl\_loss\_type & low\_var\_kl \\
entropy\_coeff & 0 \\
use\_kl\_in\_reward & False \\
n\_gpus\_per\_node & 4 \\
nnodes & 1 \\
total\_epochs & 1 \\
perceiver\_max\_prompt\_length & 2560 \\
perceiver\_max\_response\_length & 1024 \\
reasoner\_max\_prompt\_length & 8704 \\
reasoner\_max\_response\_length & 2048 \\
\bottomrule
\end{tabular}%
}
\caption{Key training hyper-parameters for PRA-GRPO.}
\label{tab:training_params}
\end{table}

\subsection{Detailed Training Setup}
\label{app:training_setup}
Table~\ref{tab:training_params} summarizes the main training hyper-parameters. We implement PRA-GRPO using the VeRL~\citep{sheng2025hybridflow} framework. The Perceiver and Reasoner share the same training configuration and differ only in the prompt and response length limits. During training, we cap the maximum image resolution at $2048 \times 32 \times 32$ pixels as the image pixel budget.

\Needspace{8\baselineskip}
\section{Evaluation Details}
\label{app:eval_details}

\subsection{Evaluation Datasets}
\label{app:eval_datasets}

We evaluate P2R on three benchmark suites used in the main results: V-Star~\citep{wu2024v-star}, HR-Bench (4K and 8K)~\citep{wang2025hr-bench}, and MME-RealWorld-Lite~\citep{zhang2024mme}. These benchmarks provide complementary evaluation settings, ranging from high-resolution fine-grained perception to broader real-world multimodal perception and reasoning.

\begin{itemize}
    \item \textbf{V-Star:} V-Star is designed to evaluate multimodal models in challenging visual scenarios where the required evidence is difficult to locate. It is built from 191 high-resolution images, with an average resolution of $2246\times1582$, and contains two sub-tasks: attribute recognition and spatial relationship reasoning. The questions are manually curated so the correct answer cannot be reliably guessed without accurate visual grounding.

    \item \textbf{HR-Bench:} HR-Bench focuses on fine-grained perception in high-resolution images. It contains two sub-tasks: Fine-grained Single-instance Perception (FSP), which evaluates recognition of detailed attributes such as color and material, and Fine-grained Cross-instance Perception (FCP), which evaluates relative position understanding across objects. Each sub-task contains 100 samples. We report results on both HR-Bench-4K and HR-Bench-8K, corresponding to cropped 4K images and original 8K images, respectively.

    \item \textbf{MME-RealWorld-Lite:} We further evaluate on MME-RealWorld-Lite, a lightweight subset of MME-RealWorld commonly used for efficient evaluation. Following the official lite setting, it contains 50 samples per task, or all samples when a task has fewer than 50 examples. The benchmark covers diverse real-world scenarios, including OCR in the wild, remote sensing, diagrams and tables, autonomous driving, and monitoring, and therefore serves as a broader test of multimodal perception and reasoning beyond the high-resolution benchmarks above.
\end{itemize}

Together, these benchmarks allow us to evaluate P2R in both fine-grained high-resolution settings and more general real-world multimodal understanding scenarios.

\subsection{Detailed Evaluation Setup}
\label{app:eval_setup}

For evaluation, we use greedy decoding with temperature 0 to ensure reproducible results. In addition, we increase the maximum image resolution to $4096 \times 32 \times 32$ pixels at evaluation time.

\Needspace{12\baselineskip}
\section{Additional Analysis}
\label{app:additional_analysis}

\subsection{Additional Training Dynamics}
\label{app:additional_training_dynamics}

\begin{figure}[H]
    \centering
    \includegraphics[width=.65\linewidth]{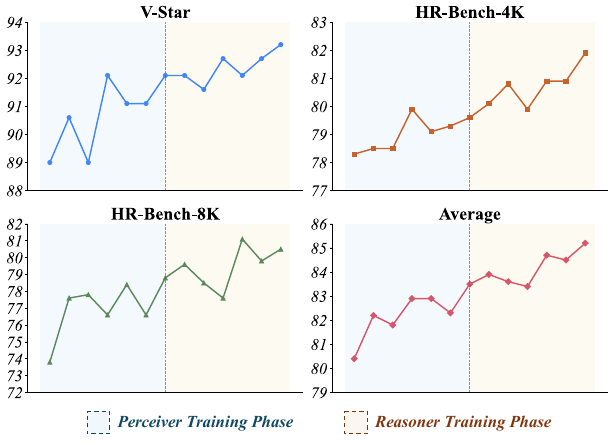}
    \caption{Evaluation accuracy dynamics on high-resolution benchmarks during the Perceiver and Reasoner training phases of PRA-GRPO.}
    \label{fig:training_dynamics_hr}
\end{figure}

\Needspace{8\baselineskip}
\paragraph{Accuracy on High-Resolution Benchmarks}
Figure~\ref{fig:training_dynamics_hr} shows the evaluation accuracy on V-Star, HR-Bench-4K, and HR-Bench-8K across the Perceiver and Reasoner training phases. We observe consistent gains on all three benchmarks, with average performance steadily improving throughout training. This result suggests that both stages of PRA-GRPO contribute to better fine-grained visual perception on challenging high-resolution images. In particular, the gains continue not only during the Perceiver phase, where the model directly learns to localize informative evidence, but also during the Reasoner phase, indicating that improving downstream reasoning can further enhance the overall perceive-to-reason pipeline.

\Needspace{8\baselineskip}
\begin{figure}[H]
    \centering
    \includegraphics[width=.65\linewidth]{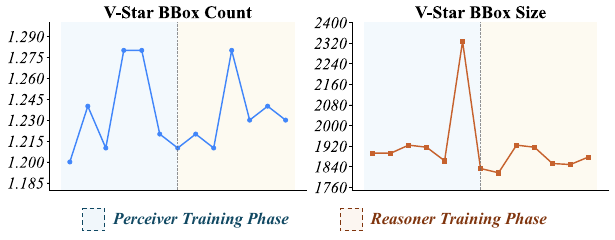}
    \caption{Dynamics of bounding box count and size on V-Star during the Perceiver and Reasoner training phases of PRA-GRPO.}
    \label{fig:training_dynamics_bbox}
\end{figure}

\paragraph{Statistics of Bounding Box Count and Size}
Figure~\ref{fig:training_dynamics_bbox} plots the average bounding box count and size on V-Star throughout training. Importantly, we do not observe a monotonic increase in either the number of predicted boxes or their spatial extent. This suggests that the model is not exploiting the reward by simply proposing more regions or enlarging boxes to cover as much of the image as possible. Instead, the model remains focused on identifying compact and informative regions that are most relevant to the final answer. At the same time, both statistics exhibit a rise-then-fall pattern, which is consistent with an exploration process: the model initially explores broader region proposals and then gradually refines them toward more selective and targeted localization. Interestingly, these box statistics also change during the Reasoner training phase. Although only the Reasoner is updated in that stage, its improvement still affects the overall role interaction in PRA-GRPO, which in turn influences the Perceiver's learned bounding-box behavior in the final pipeline.

\Needspace{28\baselineskip}
\subsection{Efficiency Analysis}
\label{app:efficiency_analysis}
\begin{figure}[H]
    \centering
    \includegraphics[width=.65\linewidth]{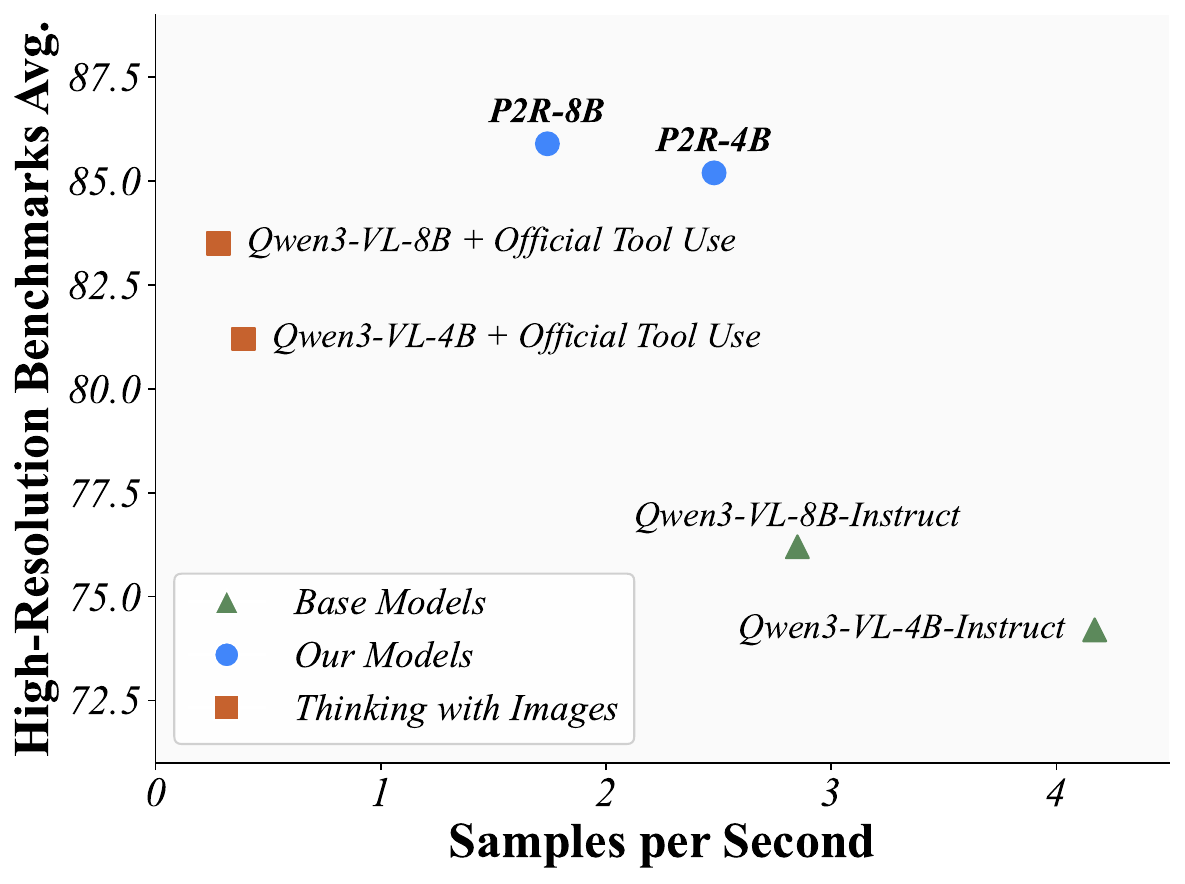}
    \caption{Comparison of inference throughput (samples per second) and average accuracy on the high-resolution benchmarks for Qwen3-VL-Instruct, P2R, and Qwen3-VL with official tool use. Official tool use denotes the released tool-calling inference pipeline of Qwen3-VL, which follows a \emph{Thinking with Images} style of interaction.}
    \label{fig:efficiency_comparison}
\end{figure}
Figure~\ref{fig:efficiency_comparison} compares throughput and average performance on the high-resolution benchmarks. Overall, P2R achieves a favorable efficiency-accuracy trade-off: compared with the corresponding Qwen3-VL-Instruct base models, P2R substantially improves benchmark performance while retaining relatively high inference efficiency. The official tool-use baseline follows the \emph{Thinking with Images} paradigm, and P2R is both more accurate and much faster than this variant. Although P2R is slower than the base models due to the additional interaction, it remains practical in terms of inference cost.

For fairness, we do not include direct efficiency comparisons with visual search methods. Our experiments use a unified vLLM~\citep{kwon2023vllm} backend, whereas visual search methods typically rely on more complex multi-stage pipelines and often use different backends. Direct wall-clock comparisons would therefore be confounded by implementation differences. Still, prior work~\citep{liu2025hide, yu2025zoom-refine} suggests that visual search methods usually incur much higher latency; for example, methods such as ZoomEye~\citep{shen2025zoomeye} are often reported to require more than $5\times$ the inference time of the underlying base model. This further suggests that P2R offers a more favorable practical efficiency-performance trade-off.

\Needspace{28\baselineskip}
\subsection{Analysis of Bounding Box Quality}
\label{app:oracle_bbox_analysis}
\noindent
\begingroup
\setlength{\intextsep}{4pt plus 1pt minus 1pt}
\begin{wrapfigure}{r}{.40\textwidth}
    \centering
    \includegraphics[width=\linewidth]{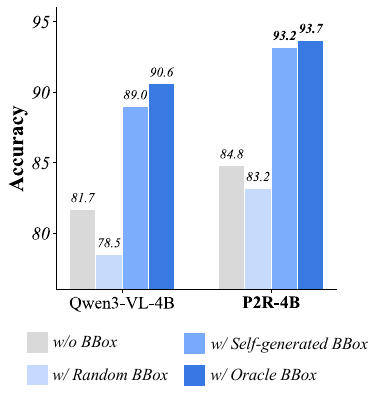}
    \caption{Bounding box inputs on V-Star for Qwen3-VL-4B and P2R-4B.}
    \label{fig:oracle_bbox_analysis}
\end{wrapfigure}
Figure~\ref{fig:oracle_bbox_analysis} highlights the importance of bounding boxes in our framework. Compared with the no-bounding-box setting, using bounding boxes leads to substantial performance gains, showing that explicitly focusing on relevant regions is critical for fine-grained visual reasoning. This confirms the importance of decoupling perception and reasoning in the P2R framework. In contrast, random bounding boxes hurt performance, further showing that accurate region localization is essential.

After PRA-GRPO training, the model's self-generated bounding boxes already achieve performance very close to that of ground-truth bounding boxes. For P2R-4B, the gap between self-generated and oracle boxes is only 0.5\% (93.2\% vs.\ 93.7\%), indicating that the learned Perceiver can accurately identify key regions without external box supervision at inference time. In addition, P2R-4B is more robust to random bounding boxes than the base model, with a much smaller degradation under noisy box inputs. This suggests that PRA-GRPO improves both region localization quality and robustness to imperfect visual hints.

\FinishSideFigure
\endgroup

\Needspace{24\baselineskip}
\subsection{Effect of Training Method}
\label{app:training_method}
\begin{figure}[H]
    \centering
    \includegraphics[width=.69\linewidth]{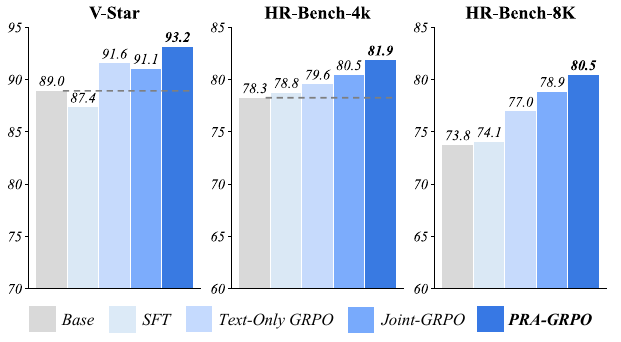}
    \caption{Comparison of training methods on Qwen3-VL-Instruct-4B with P2R inference.}
    \label{fig:ablation_training_method}
\end{figure}

Figure~\ref{fig:ablation_training_method} compares training methods under the same Qwen3-VL-Instruct-4B backbone, P2R inference, 10K training set, and two-epoch budget. SFT uses standard supervised fine-tuning; Text-Only GRPO directly optimizes answers without P2R inference during training; Joint-GRPO updates both roles without freezing either. All GRPO variants use the same binary final-answer reward. SFT yields limited and inconsistent gains, while Text-Only GRPO and Joint-GRPO improve the average score from 80.4\% to 82.7\% and 83.5\%. PRA-GRPO performs best on all three benchmarks, averaging 85.2\% and outperforming Joint-GRPO by 2.1, 1.4, and 1.6 percentage points on V-Star, HR-Bench-4K, and HR-Bench-8K, respectively. These results support the benefit of alternating role-specific optimization.

\Needspace{20\baselineskip}
\subsection{Same-Backbone Comparison}
\label{app:same_backbone_comparison}
To isolate the effect of backbone choice, we instantiate the complete P2R framework on Qwen2.5-VL-7B~\citep{bai2025qwen2}, the same backbone used by DeepEyes-7B~\citep{zheng2025deepeyes}. As shown in Table~\ref{tab:same_backbone_comparison}, P2R-7B improves the average score from 72.7\% to 79.7\%, matches DeepEyes-7B on V-Star, and performs better on both HR-Bench settings. It also achieves a higher average score than DeepEyes-7B and ZoomEye-7B~\citep{shen2025zoomeye}. These results show that the gains of P2R are not specific to the Qwen3-VL backbone.

\begin{table}[H]
\centering
\small
\paperfit{
\renewcommand{\arraystretch}{1.15}
\begin{tabular}{lcccc}
\toprule
\textbf{Method} & \textbf{V-Star} & \textbf{HR-4K} & \textbf{HR-8K} & \textbf{Avg.} \\
\midrule
Qwen2.5-VL-7B & 78.5 & 71.6 & 67.9 & 72.7 \\
DeepEyes-7B & 90.1 & 75.1 & 72.6 & 79.3 \\
ZoomEye-7B & \textbf{90.6} & 69.6 & 69.3 & 76.5 \\
P2R-7B & 90.1 & \textbf{75.4} & \textbf{73.6} & \textbf{79.7} \\
\bottomrule
\end{tabular}
}
\caption{Same-backbone comparison on Qwen2.5-VL-7B. P2R-7B denotes the complete model with PRA-GRPO training and P2R inference.}
\label{tab:same_backbone_comparison}
\end{table}

\subsection{GRPO vs. DAPO}
\label{app:grpo_vs_dapo}
DAPO~\citep{yu2026dapo} is an improved variant of GRPO~\citep{schulman2017ppo, liu2025dr-grpo, yan2026luffy, zheng2025gspo, fu2025srft} that has shown strong performance on mathematical reasoning tasks. Compared with standard GRPO, DAPO introduces several modifications, including removing the KL divergence term, clip-higher, dynamic sampling, token-level policy gradient loss, and overlong reward shaping. Prior work~\citep{tang2025rethinking,liu2026gdpo} has shown that these changes can lead to clear gains over GRPO on text-based reasoning benchmarks, and recent work~\citep{lai2025mini-o3, wei2026zooming} on fine-grained visual perception has also adopted this training recipe.

However, in our setting, we find that DAPO is less suitable for training the Perceiver. In particular, during the Perceiver phase, DAPO causes the predicted bounding box count to drop rapidly and quickly converge to one. This behavior is problematic for tasks that require multiple boxes. For example, many samples in the spatial relationship reasoning split of V-Star require comparing the relative positions of two objects. If the model outputs only a single bounding box, it cannot reliably capture both objects, leading to a clear performance drop. On the 4B model, DAPO achieves only 85.8\% on V-Star Spatial, whereas GRPO reaches 94.7\%.

We hypothesize that this issue is related to the removal of the KL divergence term in DAPO. Without KL regularization, the policy can more easily drift away from the original response pattern and collapse to a simpler mode that generates only one bounding box. While such behavior may not be problematic in pure text reasoning, it is harmful in our setting, where the model must maintain flexible multi-box perception behavior. We therefore use GRPO instead of DAPO in all main experiments.

\subsection{More Cases}
\label{app:more_cases}
\begin{figure}[H]
    \centering
    \includegraphics[width=.82\linewidth,height=.82\textheight,keepaspectratio]{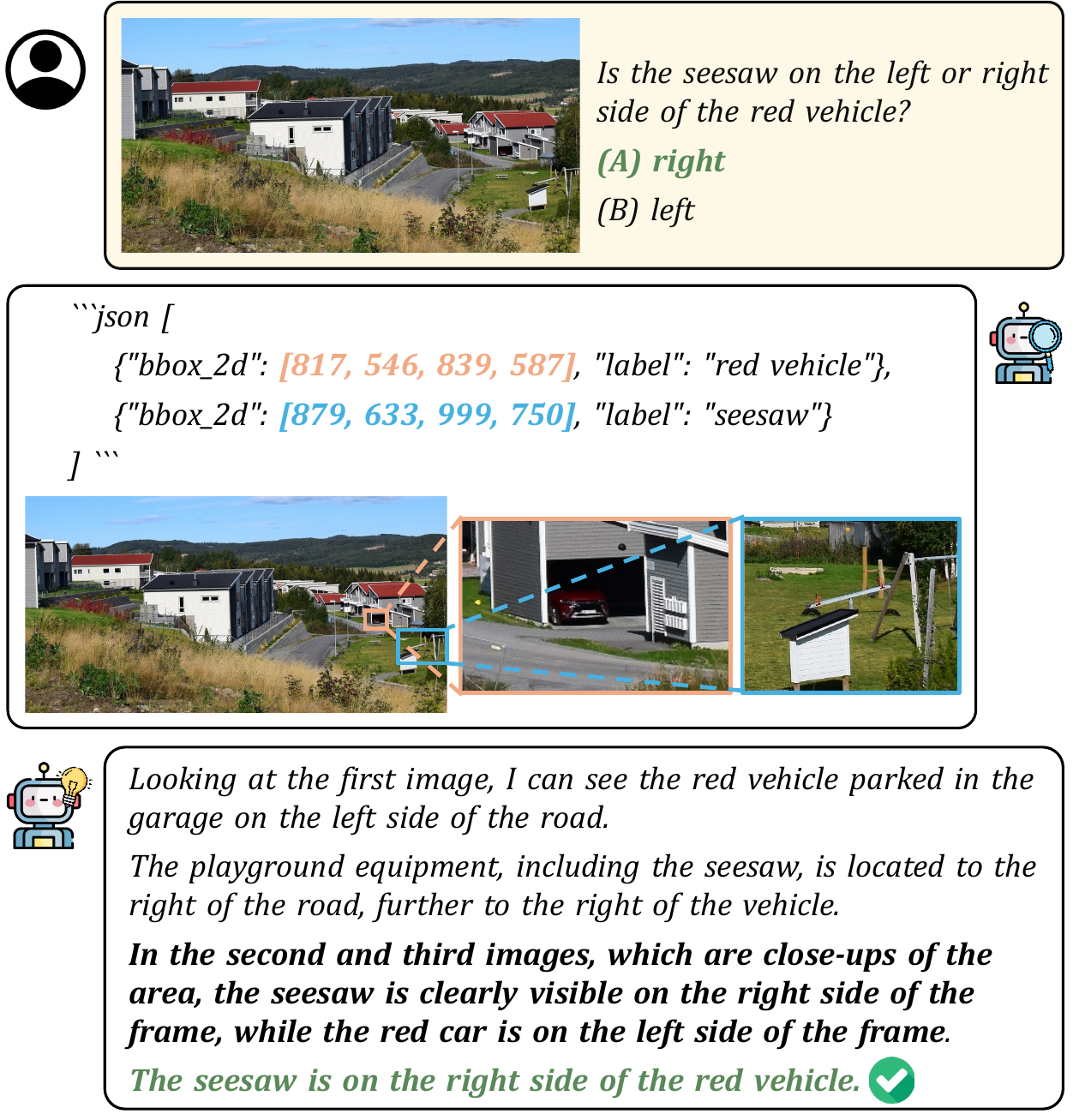}
    \caption{A successful case of P2R-4B on the V-Star Spatial.}
    \label{fig:good_case_vstar_spatial}
\end{figure}

\begin{figure}[H]
    \centering
    \includegraphics[width=.82\linewidth,height=.82\textheight,keepaspectratio]{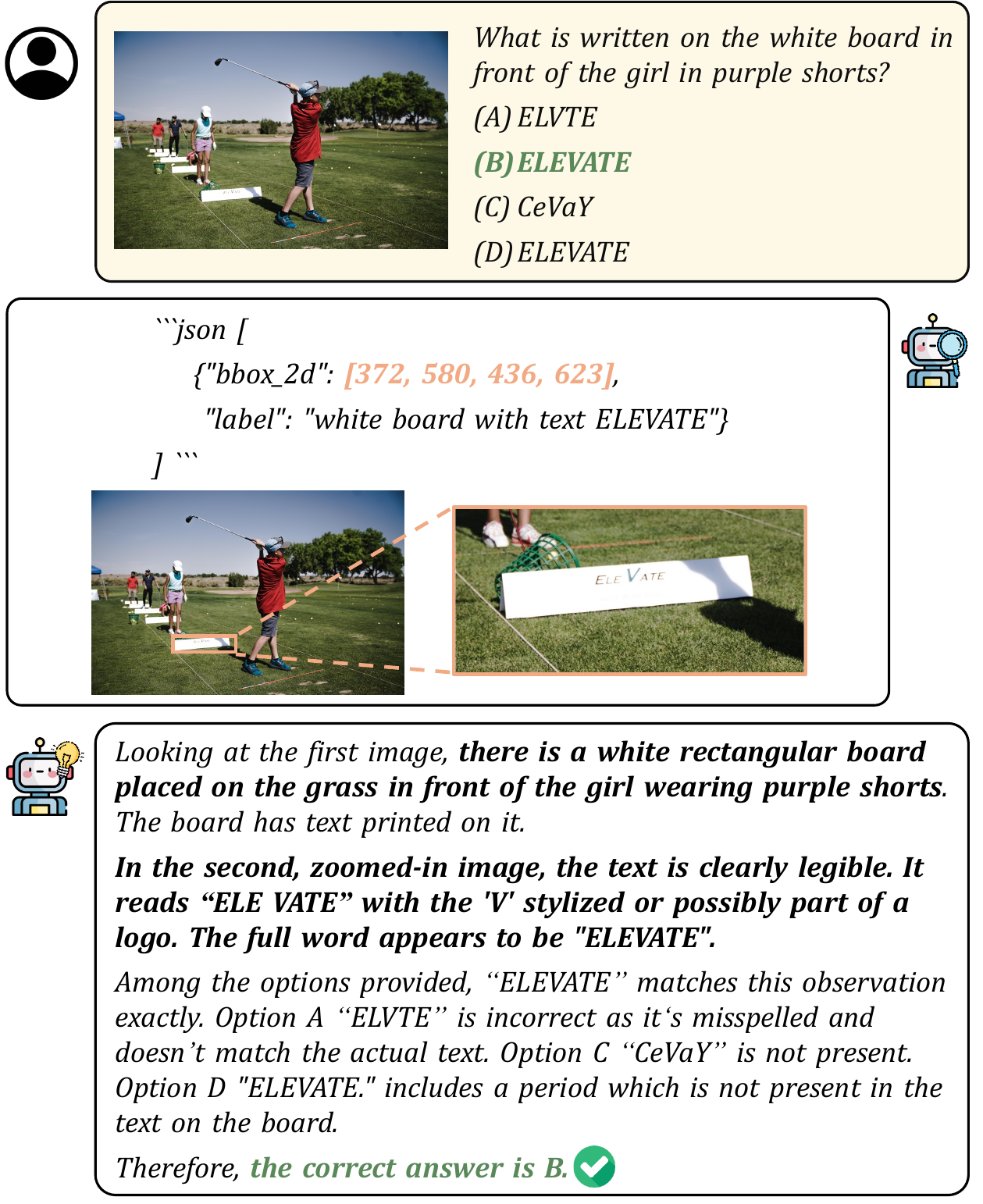}
    \caption{A successful case of P2R-4B on the HR-Bench FSP.}
    \label{fig:good_case_hrbench_fsp}
\end{figure}

\begin{figure}[H]
    \centering
    \includegraphics[width=.82\linewidth,height=.82\textheight,keepaspectratio]{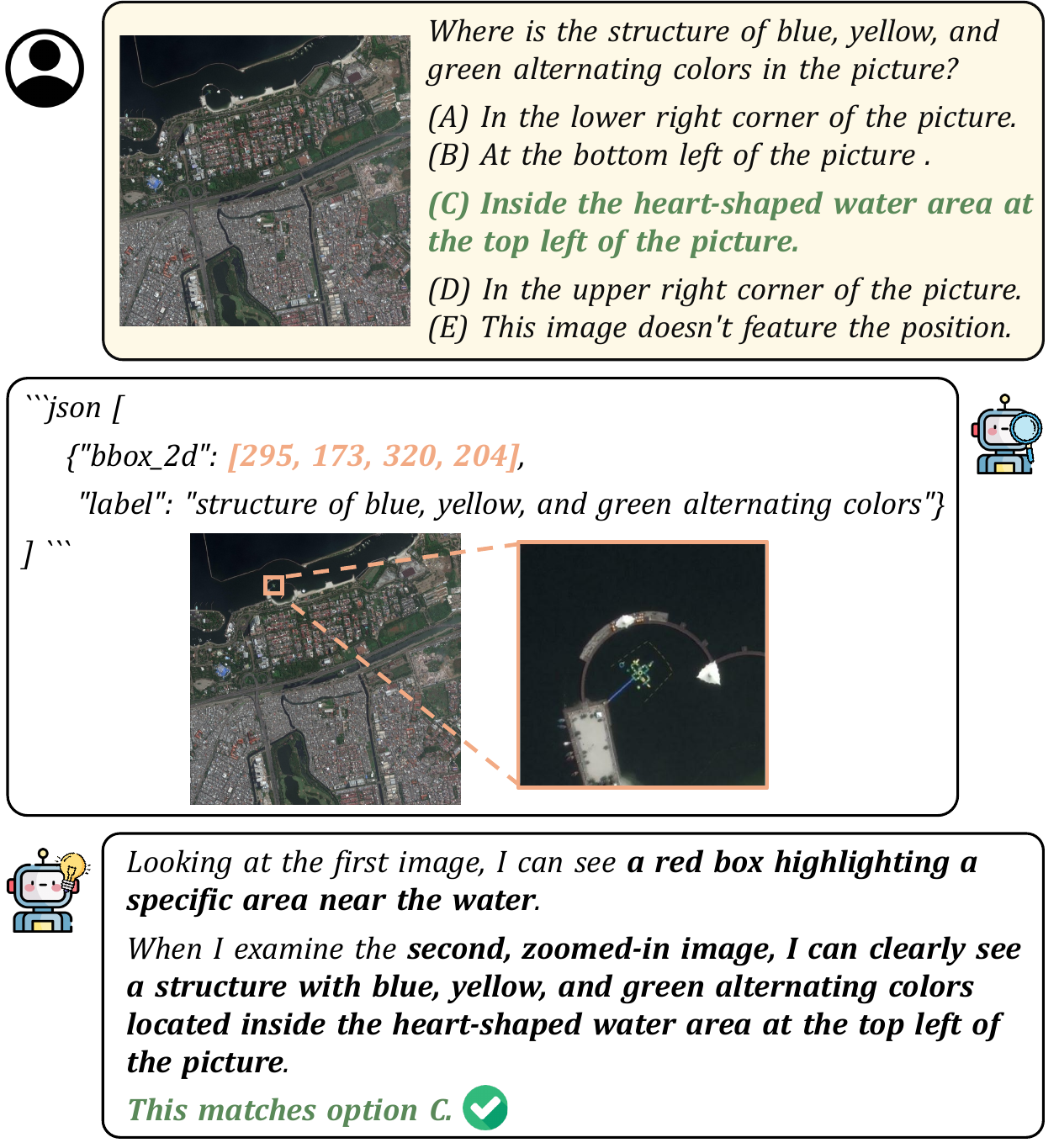}
    \caption{A successful case of P2R-4B on the MME-RealWorld-Lite Perception Remote Sensing.}
    \label{fig:good_case_mme_pr}
\end{figure}

\begin{figure}[H]
    \centering
    \includegraphics[width=.82\linewidth,height=.82\textheight,keepaspectratio]{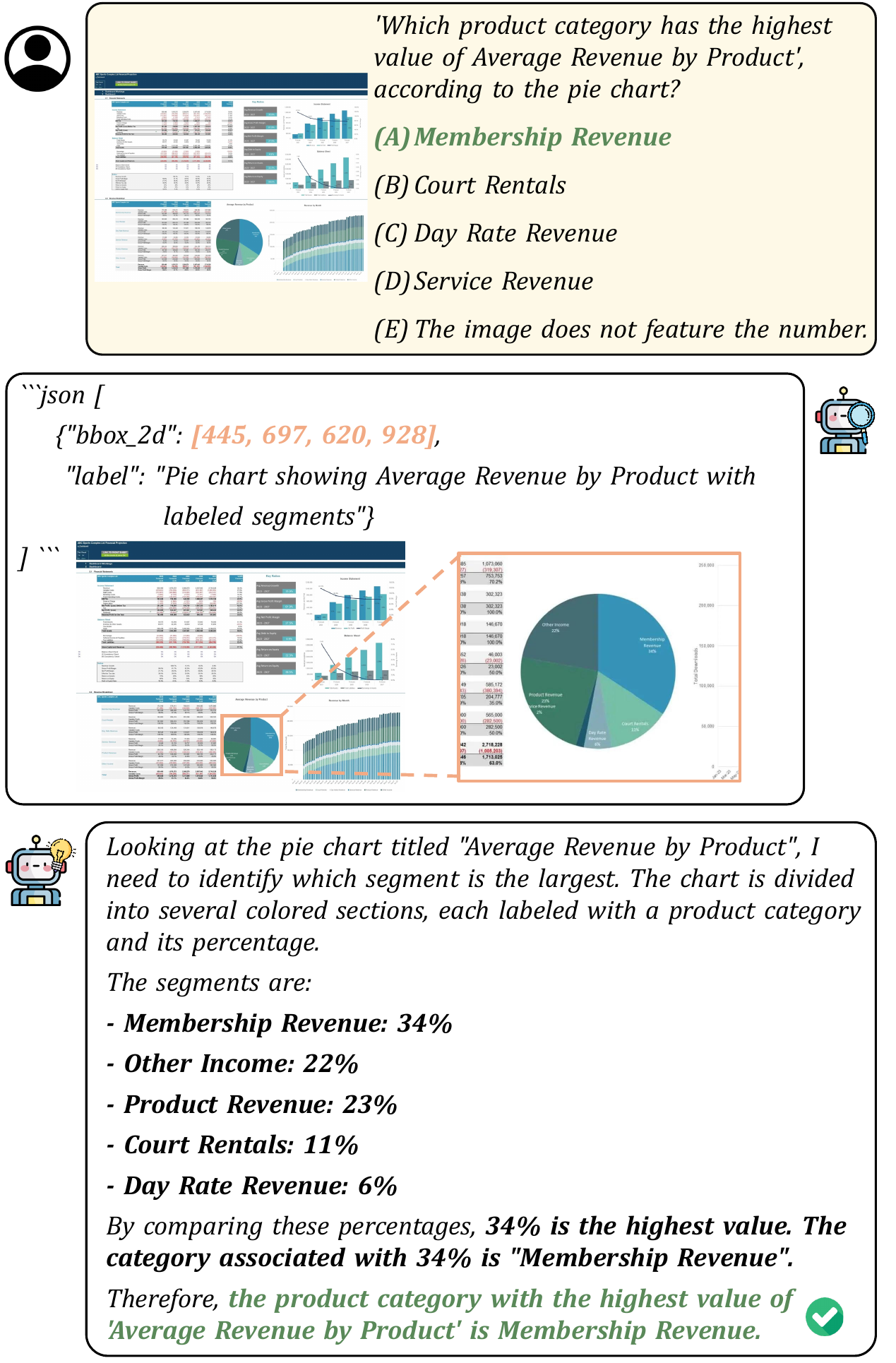}
    \caption{A successful case of P2R-4B on the MME-RealWorld-Lite Reasoning Diagram and Table.}
    \label{fig:good_case_mme_rd}
\end{figure}

\begin{figure}[H]
    \centering
    \includegraphics[width=.82\linewidth,height=.82\textheight,keepaspectratio]{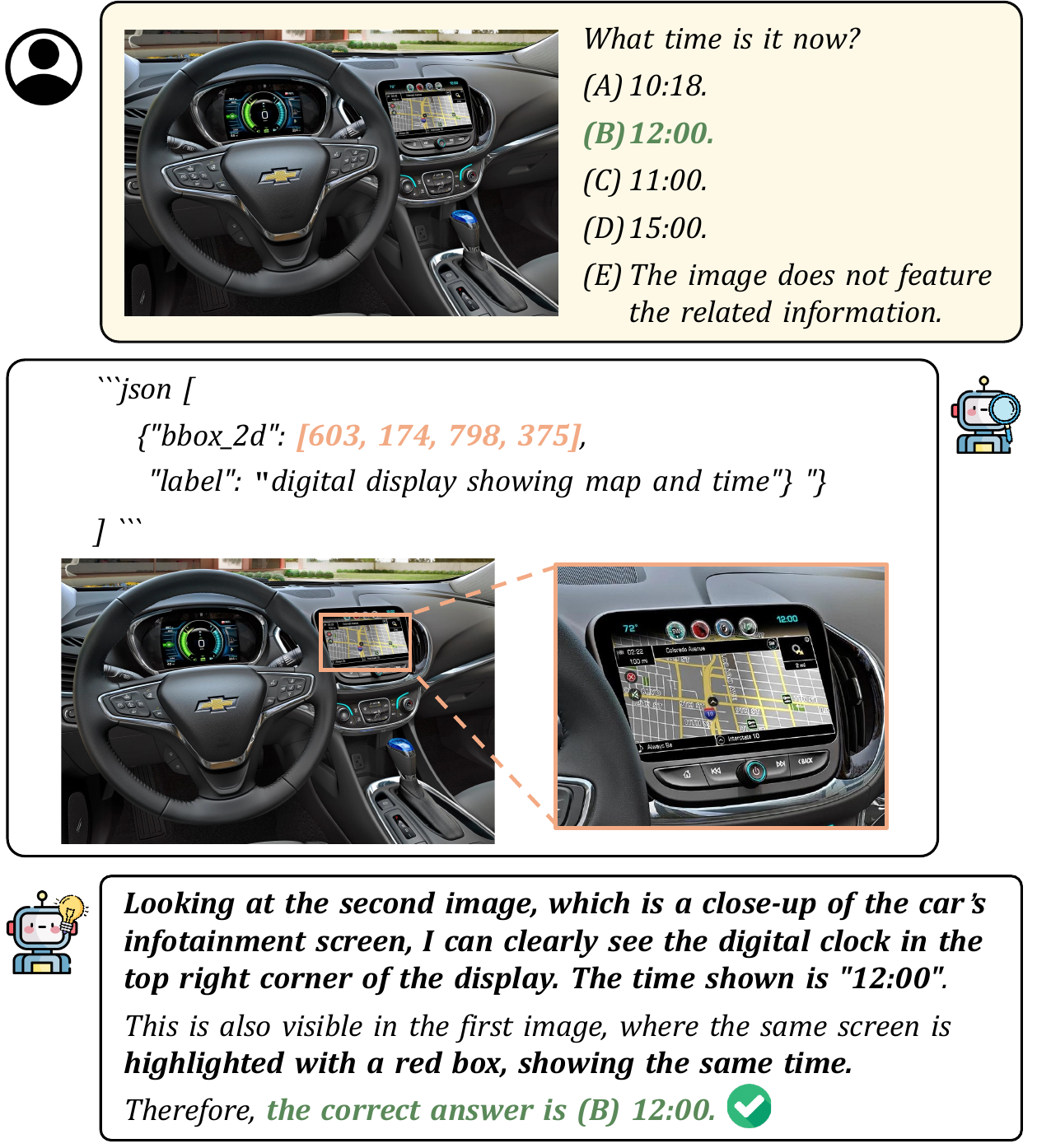}
    \caption{A successful case of P2R-4B on the MME-RealWorld-Lite Reasoning OCR with Complex Context.}
    \label{fig:good_case_mme_ro}
\end{figure}

Figures~\ref{fig:good_case_vstar_attr}, \ref{fig:good_case_vstar_spatial}, \ref{fig:good_case_hrbench_fsp}, \ref{fig:good_case_mme_pr}, \ref{fig:good_case_mme_rd}, and \ref{fig:good_case_mme_ro} show additional successful cases of P2R-4B across diverse benchmarks and task types, including fine-grained attribute recognition, spatial relation reasoning, chart understanding, remote sensing perception, and OCR-intensive reasoning. Overall, these examples exhibit a consistent perceive-to-reason pattern: the model first identifies a compact region relevant to the query, and then uses the zoomed-in crop to extract fine-grained evidence for the final answer. This behavior is particularly useful when the target evidence is small, subtle, or embedded in cluttered high-resolution scenes. Across these cases, P2R-4B can successfully localize and reason over key visual evidence such as small symbols, distant objects, fine-grained text, chart segments, and tiny structures in remote-sensing images.

\begin{figure}[H]
    \centering
    \includegraphics[width=.82\linewidth,height=.82\textheight,keepaspectratio]{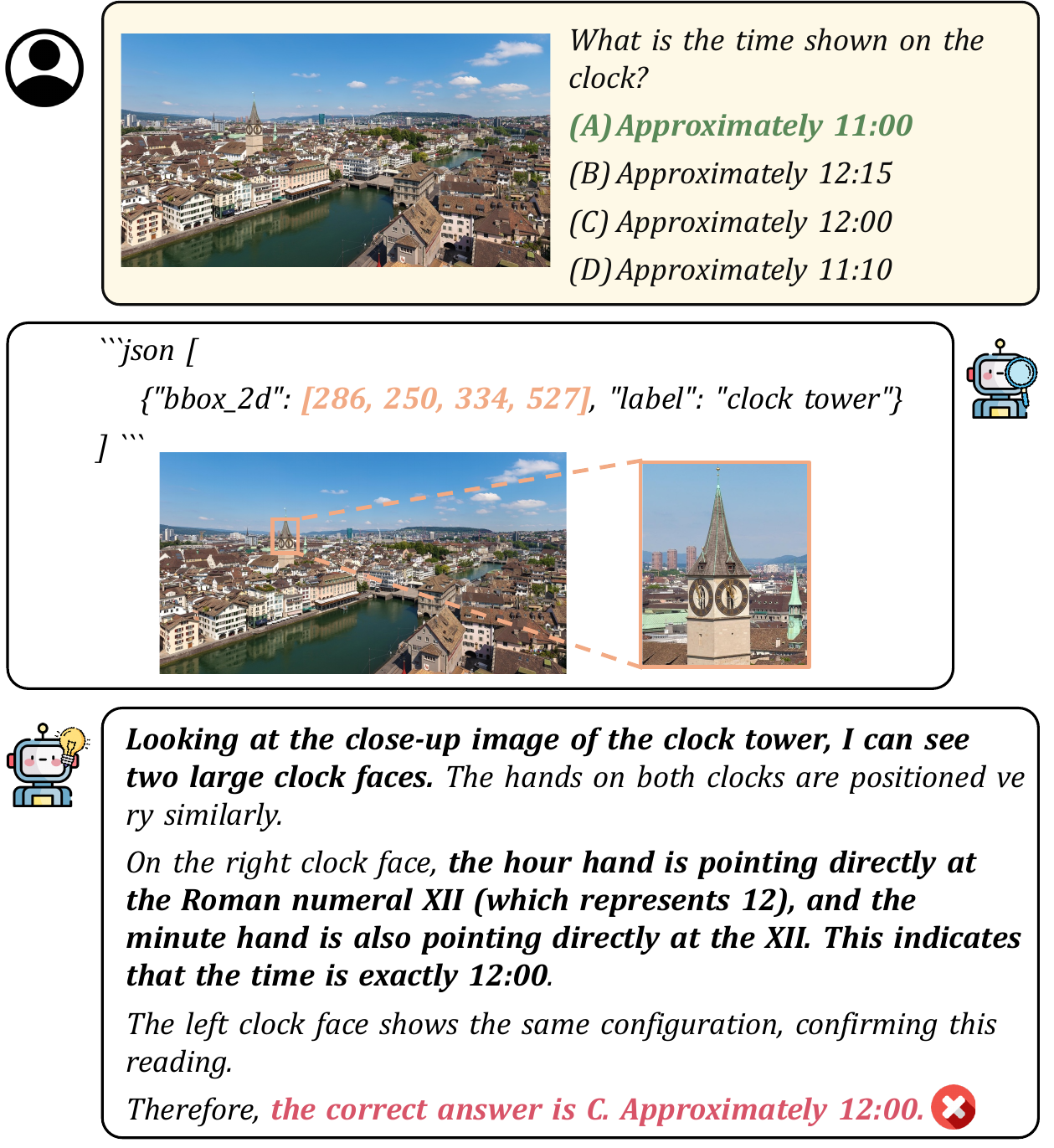}
    \caption{A failure case of P2R-4B on the HR-Bench FSP.}
    \label{fig:bad_case_hrbench_fsp}
\end{figure}

\begin{figure}[H]
    \centering
    \includegraphics[width=.82\linewidth,height=.82\textheight,keepaspectratio]{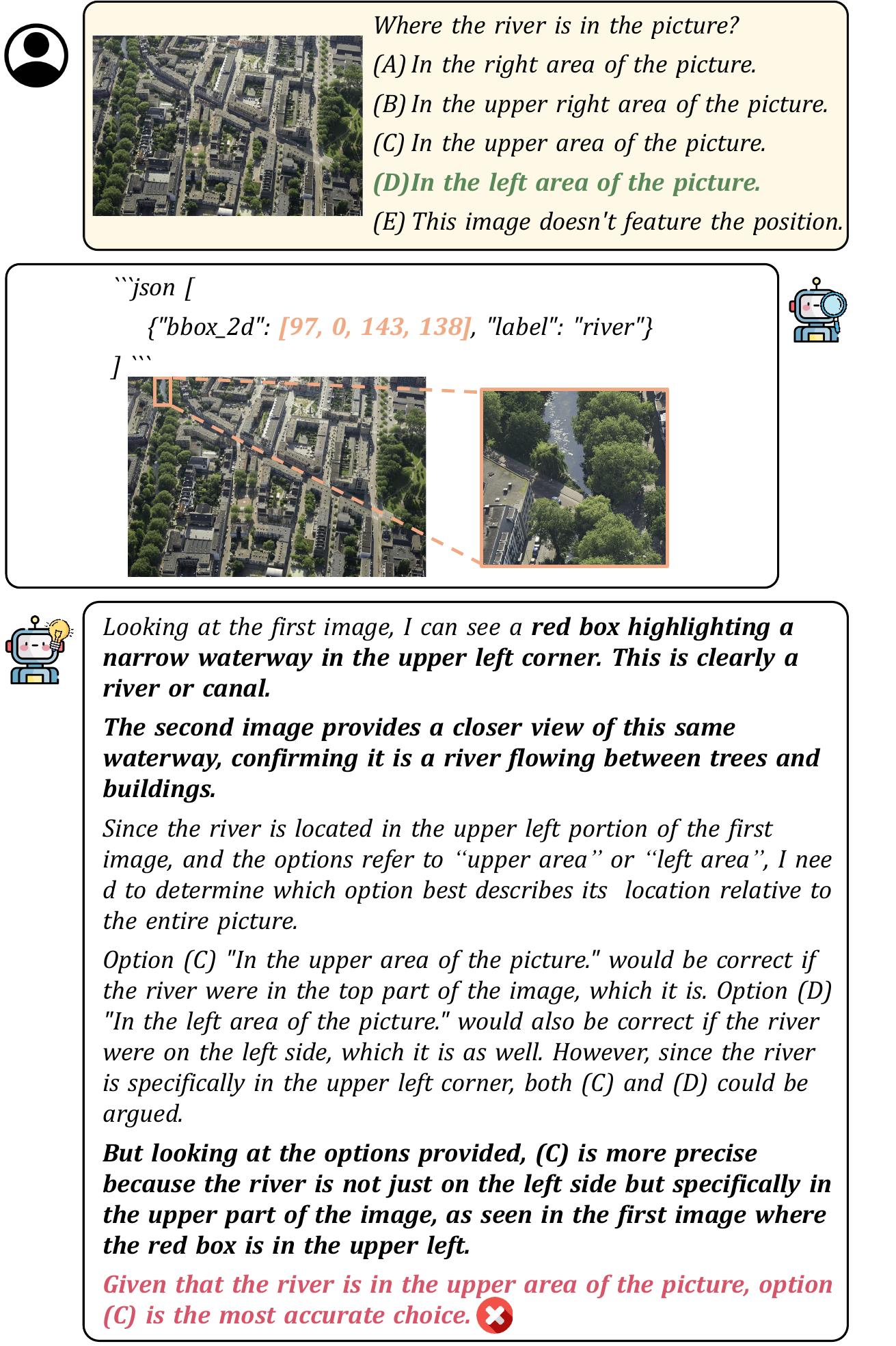}
    \caption{A failure case of P2R-4B on the MME-RealWorld-Lite Perception Remote Sensing.}
    \label{fig:bad_case_mme_pr}
\end{figure}

We also show two representative failure cases in Figures~\ref{fig:bad_case_hrbench_fsp} and~\ref{fig:bad_case_mme_pr}. In Figure~\ref{fig:bad_case_hrbench_fsp}, the model correctly localizes the clock tower and zooms into the relevant clock face, indicating that the Perceiver identifies the right evidence. The error instead comes from fine-grained visual recognition in the cropped region: the model mistakenly interprets the hour hand as pointing to 12 rather than 11, which leads to the wrong answer. This example suggests that even with accurate localization, P2R-4B can still fail on subtle visual reading tasks that require precise interpretation of small details.

In Figure~\ref{fig:bad_case_mme_pr}, the model also successfully detects the river, but the selected box covers only part of the river rather than the full river region on the left side of the image. As a result, the cropped evidence does not fully preserve the spatial context needed to distinguish whether the answer should be ``left'' or ``upper.'' This incomplete localization introduces ambiguity in the reasoning stage. If the full left-side river region were captured, this ambiguity would likely be avoided. This case highlights that, beyond finding the relevant object, the spatial extent of the selected region is also crucial for correct downstream reasoning.

\begingroup
\setlength{\intextsep}{26pt plus 2pt minus 2pt}
\begin{figure}[H]
    \centering
    \includegraphics[width=.82\linewidth,height=.82\textheight,keepaspectratio]{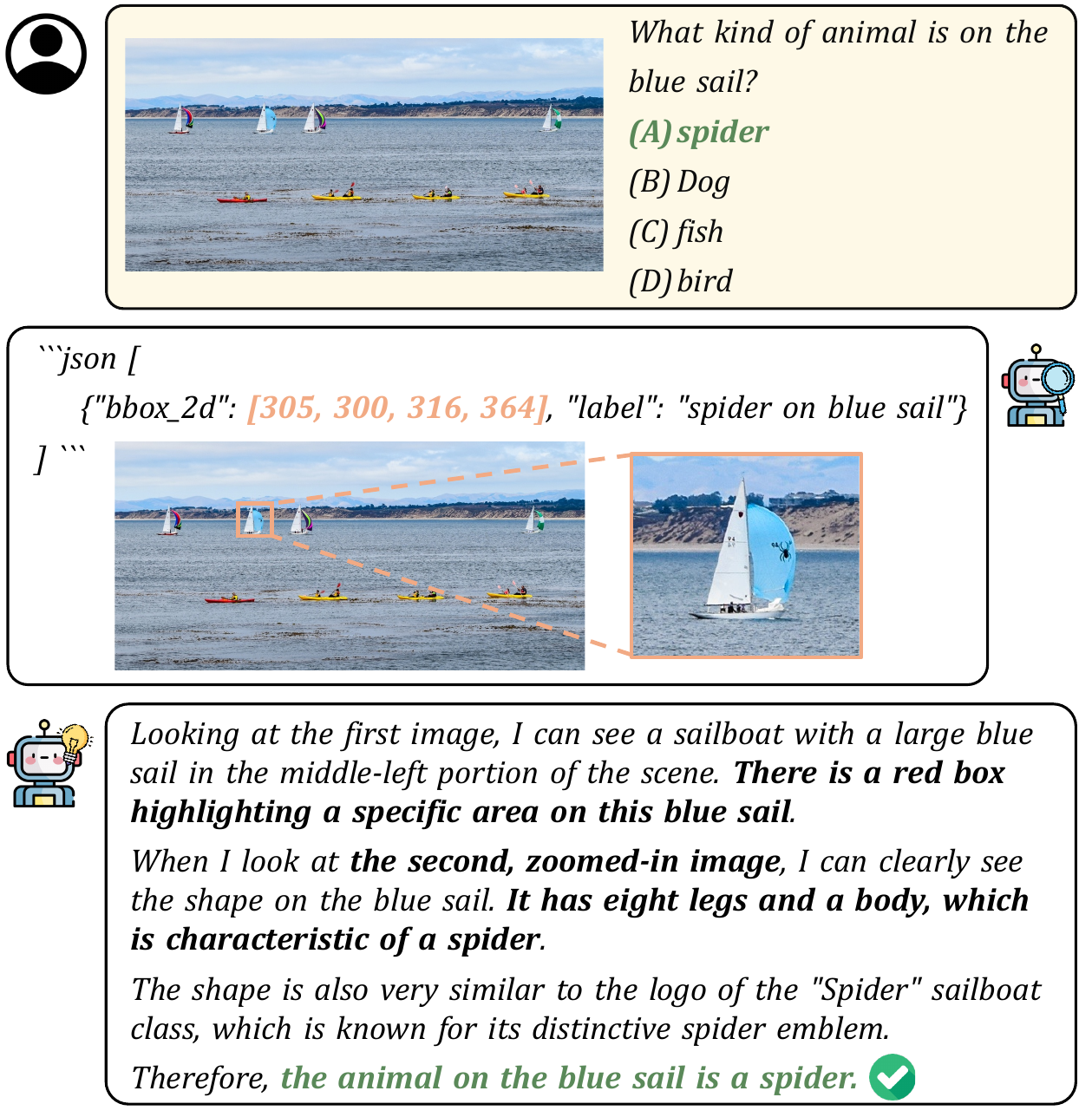}
    \caption{A successful case of P2R-4B on the V-Star Attribute.}
    \label{fig:good_case_vstar_attr}
\end{figure}
\endgroup

\end{document}